\documentclass{article}

    \PassOptionsToPackage{numbers, compress}{natbib}

 \usepackage[eandd,preprint]{neurips_2026}

\usepackage[utf8]{inputenc} 
\usepackage[T1]{fontenc}    
\usepackage{hyperref}       
\usepackage{url}            
\usepackage{longtable}
\usepackage{booktabs}       
\usepackage{amsfonts}       
\usepackage{nicefrac}       
\usepackage{microtype}      

\usepackage[most]{tcolorbox}  
\usepackage{amsmath}    
\usepackage[table]{xcolor}
\usepackage{tikz}   
\usepackage{enumitem} 
\usepackage{caption}
\usepackage{bbm}
\usepackage{multirow}
\usepackage{listings}
\usepackage{amssymb}
\usepackage{wrapfig}
\usepackage{fontawesome5}

\usepackage{tocbibind}  
\usepackage{titletoc}   

\definecolor{casebg}{HTML}{F7F7FF}      
\definecolor{casetitlebg}{HTML}{8F8FE0} 
\newcommand{\tcolorboxCase}[2]{%
\begin{tcolorbox}[
    title=#1,
    colback=casebg, 
    colframe=casetitlebg,
    rounded corners,
    sharp corners=northeast,
    sharp corners=southwest,
    width=1.00\linewidth,
    boxsep=2pt,
    top=2pt,
    bottom=2pt,
    enhanced,
    breakable,
    before=\setlength{\parindent}{0pt},
]

#2
\end{tcolorbox}
}

\newcommand{\tcolorboxPrompt}[2]{%
\begin{tcolorbox}[
    title=#1,
    colback=cyan!3, 
    colframe=blue!50!black!60,
    rounded corners,
    sharp corners=northeast,
    sharp corners=southwest,
    width=1.00\linewidth,
    boxsep=2pt,
    top=2pt,
    bottom=2pt,
    enhanced,
    breakable,
    before=\setlength{\parindent}{0pt},
]

#2
\end{tcolorbox}
}

\definecolor{codebg}{RGB}{248,248,248}
\definecolor{codeframe}{RGB}{220,220,220}
\definecolor{codecomment}{RGB}{100,120,100}
\definecolor{codekeyword}{RGB}{40,80,140}
\definecolor{codestring}{RGB}{150,60,60}

\lstdefinestyle{icmlcode}{
  backgroundcolor=\color{codebg},
  frame=single,
  rulecolor=\color{codeframe},
  basicstyle=\ttfamily\footnotesize,
  columns=fullflexible,
  keepspaces=true,
  breaklines=true,
  breakatwhitespace=false,
  tabsize=2,
  showstringspaces=false,
  numbers=none,
  numberstyle=\tiny\color{gray},
  numbersep=6pt,
  xleftmargin=0em,
  framexleftmargin=0.5em,
  aboveskip=0.75em,
  belowskip=0.75em,
  captionpos=b,
  keywordstyle=\color{codekeyword}\bfseries,
  commentstyle=\color{codecomment}\itshape,
  stringstyle=\color{codestring}
}

\lstdefinestyle{python}{
  style=icmlcode,
  language=Python
}

\lstdefinelanguage{json}{
  basicstyle=\ttfamily\footnotesize,
  showstringspaces=false,
  breaklines=true,
  morestring=[b]",
  stringstyle=\color{codestring},
  literate=
   *{0}{{{\color{codekeyword}0}}}{1}
    {1}{{{\color{codekeyword}1}}}{1}
    {2}{{{\color{codekeyword}2}}}{1}
    {3}{{{\color{codekeyword}3}}}{1}
    {4}{{{\color{codekeyword}4}}}{1}
    {5}{{{\color{codekeyword}5}}}{1}
    {6}{{{\color{codekeyword}6}}}{1}
    {7}{{{\color{codekeyword}7}}}{1}
    {8}{{{\color{codekeyword}8}}}{1}
    {9}{{{\color{codekeyword}9}}}{1}
    {:}{{{\color{gray}{:}}}}{1}
    {,}{{{\color{gray}{,}}}}{1}
    {\{}{{{\color{gray}{\{}}}}{1}
    {\}}{{{\color{gray}{\}}}}}{1}
    {[}{{{\color{gray}{[}}}}{1}
    {]}{{{\color{gray}{]}}}}{1},
}

\lstdefinestyle{json}{
  style=icmlcode,
  language=json
}

\lstdefinelanguage{yaml}{
  keywords={true,false,null,y,n},
  keywordstyle=\color{codekeyword}\bfseries,
  basicstyle=\ttfamily\footnotesize,
  sensitive=false,
  comment=[l]{\#},
  commentstyle=\color{codecomment}\itshape,
  morestring=[b]",
  stringstyle=\color{codestring},
  showstringspaces=false,
  breaklines=true,
  literate=
    {---}{{{\color{gray}{---}}}}{3}
    {...}{{{\color{gray}{...}}}}{3}
    {:}{{{\color{gray}{:}}}}{1}
    {-}{{{\color{gray}{-}}}}{1}
    {>}{{{\color{gray}{>}}}}{1}
    {|}{{{\color{gray}{|}}}}{1},
}

\lstdefinestyle{yaml}{
  style=icmlcode,
  language=yaml
}

\definecolor{tcbcodekeyword}{RGB}{40,80,140}
\definecolor{tcbcodecomment}{RGB}{90,120,90}
\definecolor{tcbcodestring}{RGB}{150,70,70}

\lstdefinelanguage{json}{
  morestring=[b]",
  showstringspaces=false,
  literate=
   *{0}{{{\color{tcbcodekeyword}0}}}{1}
    {1}{{{\color{tcbcodekeyword}1}}}{1}
    {2}{{{\color{tcbcodekeyword}2}}}{1}
    {3}{{{\color{tcbcodekeyword}3}}}{1}
    {4}{{{\color{tcbcodekeyword}4}}}{1}
    {5}{{{\color{tcbcodekeyword}5}}}{1}
    {6}{{{\color{tcbcodekeyword}6}}}{1}
    {7}{{{\color{tcbcodekeyword}7}}}{1}
    {8}{{{\color{tcbcodekeyword}8}}}{1}
    {9}{{{\color{tcbcodekeyword}9}}}{1}
    {:}{{{\color{gray}{:}}}}{1}
    {,}{{{\color{gray}{,}}}}{1}
    {\{}{{{\color{gray}{\{}}}}{1}
    {\}}{{{\color{gray}{\}}}}}{1}
    {[}{{{\color{gray}{[}}}}{1}
    {]}{{{\color{gray}{]}}}}{1},
}

\lstdefinelanguage{tcbyaml}{
  keywords={true,false,null,y,n},
  keywordstyle=\color{tcbcodekeyword}\bfseries,
  basicstyle=\ttfamily\footnotesize,
  sensitive=false,
  comment=[l]{\#},
  commentstyle=\color{tcbcodecomment}\itshape,
  morestring=[b]",
  stringstyle=\color{tcbcodestring},
  showstringspaces=false,
  breaklines=true,
  literate=
    {---}{{{\color{gray}{---}}}}{3}
    {...}{{{\color{gray}{...}}}}{3}
    {:}{{{\color{gray}{:}}}}{1}
    {-}{{{\color{gray}{-}}}}{1}
    {>}{{{\color{gray}{>}}}}{1}
    {|}{{{\color{gray}{|}}}}{1},
}

\lstdefinestyle{icmltcblistingcode}{
  basicstyle=\ttfamily\footnotesize,
  columns=fullflexible,
  keepspaces=true,
  breaklines=true,
  breakatwhitespace=false,
  tabsize=2,
  showstringspaces=false,
  frame=none,
  numbers=none,
  xleftmargin=0em,
  framexleftmargin=0.5em,
  numberstyle=\tiny\color{gray},
  numbersep=6pt,
  keywordstyle=\color{tcbcodekeyword}\bfseries,
  commentstyle=\color{tcbcodecomment}\itshape,
  stringstyle=\color{tcbcodestring}
}

\newtcblisting{tcbCodeListing}[3][]{%
  title={#2},
  listing only,
  listing options={
    style=icmltcblistingcode,
    language={#3},
    {#1}
  },
  colback=cyan!3,
  colframe=blue!50!black!60,
  rounded corners,
  sharp corners=northeast,
  sharp corners=southwest,
  width=1.00\linewidth,
  boxsep=2pt,
  top=2pt,
  bottom=2pt,
  left=2pt,
  right=2pt,
  enhanced,
  breakable,
  before=\setlength{\parindent}{0pt},
}

\newcommand{\bench}{\textsc{$\pi$-Bench}}

\newcommand{\statusbox}[2]{%
  {\setlength{\fboxsep}{0.8pt}%
   \raisebox{0pt}[0pt][0pt]{\colorbox{#1}{\textit{#2}}}}%
}

\newcommand{\completed}{\statusbox{green!20}{completed}}
\newcommand{\inferred}{\statusbox{cyan!20}{inferred}}
\newcommand{\provided}{\statusbox{red!20}{provided}}

\newcommand{\intentbox}[2]{%
  {\setlength{\fboxsep}{0.8pt}%
   \raisebox{0pt}[0pt][0pt]{{$#2$}}}%
}
\newcommand{\intentcompleted}{\intentbox{green!20}{\mathcal{I}_{\text{com}}}}
\newcommand{\intentinferred}{\intentbox{cyan!20}{\mathcal{I}_{\text{inf}}}}
\newcommand{\intentprovided}{\intentbox{red!20}{\mathcal{I}_{\text{pro}}}}

\newcommand{\projectpage}{\raisebox{-1.5pt}{\faIcon{globe}}}
\newcommand{\github}{\raisebox{-1.5pt}{\includegraphics[height=1.05em]{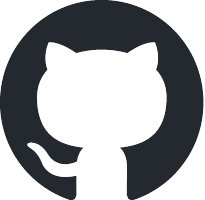}}}

\NewDocumentCommand{\yafu}
{ mO{} }{\textcolor{blue}{\textsuperscript{\textit{yafu}}\textsf{\textbf{\small[#1]}}}}

\NewDocumentCommand{\haodi}
{ mO{} }{\textcolor{blue}{\textsuperscript{\textit{haodi}}\textsf{\textbf{\small[#1]}}}}

\NewDocumentCommand{\shunkai}
{ mO{} }{\textcolor{blue}{\textsuperscript{\textit{shunkai}}\textsf{\textbf{\small[#1]}}}}

\NewDocumentCommand{\luxin}
{ mO{} }{\textcolor{blue}{\textsuperscript{\textit{luxin}}\textsf{\textbf{\small[#1]}}}}

\NewDocumentCommand{\runquan}
{ mO{} }{\textcolor{blue}{\textsuperscript{\textit{runquan}}\textsf{\textbf{\small[#1]}}}}

\NewDocumentCommand{\yangyang}
{ mO{} }{\textcolor{blue}{\textsuperscript{\textit{yangyang}}\textsf{\textbf{\small[#1]}}}}

\title{\textsc{$\pi$-Bench}: Evaluating Proactive Personal \\ Assistant Agents in Long-Horizon Workflows}

\author{
\begin{minipage}{\textwidth}
\centering
\normalfont
\textbf{Haoran Zhang}\textsuperscript{1,2,*}\hspace{1em}
\textbf{Luxin Xu}\textsuperscript{3,2,*}\hspace{1em}
\textbf{Zhilin Wang}\textsuperscript{4,2,*}\hspace{1em}
\textbf{Runquan Gui}\textsuperscript{4,2,*}\\
\textbf{Shunkai Zhang}\textsuperscript{5,2}\hspace{1em}
\textbf{Haodi Lei}\textsuperscript{6,2}\hspace{1em}
\textbf{Zihao He}\textsuperscript{7}\hspace{1em}
\textbf{Bingsu He}\textsuperscript{8}\hspace{1em}
\textbf{Chicheng Qin}\textsuperscript{5}\\
\textbf{Tong Zhu}\textsuperscript{9,2}\hspace{1em}
\textbf{Xiaoye Qu}\textsuperscript{2}\hspace{1em}
\textbf{Yang Yang}\textsuperscript{1,$\dagger$}\hspace{1em}
\textbf{Yu Cheng}\textsuperscript{10,2,$\dagger$}\hspace{1em}
\textbf{Yafu Li}\textsuperscript{2,10,$\dagger$}
\\[0.6em]
{\small
\textsuperscript{1} Shanghai Jiao Tong University \quad
\textsuperscript{2} Shanghai AI Laboratory \quad
\textsuperscript{3} Fudan University \\
\textsuperscript{4} University of Science and Technology of China \quad
\textsuperscript{5} Peking University \quad
\textsuperscript{6} Nanjing University \\
\textsuperscript{7} Zhejiang University \quad
\textsuperscript{8} Tongji University  \quad
\textsuperscript{9} Soochow University \\
\textsuperscript{10} The Chinese University of Hong Kong
}
\end{minipage}
}

\begin{document}

\maketitle

\renewcommand{\thefootnote}{\fnsymbol{footnote}}
\footnotetext[1]{Equal contribution.}
\footnotetext[2]{Corresponding authors: Yafu Li \texttt{<yafuly@gmail.com>}, Yu Cheng \texttt{<chengyu@cse.cuhk.edu.hk>}, Yang Yang \texttt{<angelayang@sjtu.edu.cn>}.}

\begin{abstract}
The rise of personal assistant agents, e.g., OpenClaw, highlights the growing potential of large language models to support users across everyday life and work. A core challenge in these settings is proactive assistance, since users often begin with underspecified requests and leave important needs, constraints, or preferences unstated. 
However, existing benchmarks rarely evaluate whether agents can identify and act on such hidden intents before they are explicitly stated, especially in sustained multi-turn interactions where user needs emerge gradually.
To address this gap, we introduce {\bench}, a benchmark for proactive assistance comprising 100 multi-turn tasks across 5 domain-specific user personas. By incorporating hidden user intents, inter-task dependencies, and cross-session continuity, {\bench} evaluates agents’ ability to anticipate and address user needs over extended interactions, jointly measuring proactivity and task completion in long-horizon trajectories that better reflect real-world use.
Experiments show (1) proactive assistance remains challenging, (2) a clear distinction between task completion and proactivity, and (3) the value of prior interaction for proactive intent resolution in later tasks.
\end{abstract}


\begin{center}
\vspace{-1em}
~\projectpage~\href{http://simplified-reasoning.github.io/Pi-Bench}{{\text{Project Page}}}
\quad \quad \quad
~\github~\href{https://github.com/Simplified-Reasoning/Pi-Bench}{{\text{Code}}}
\end{center}

\section{Introduction}
\label{sec:intro}

The emergence of personal assistant agents such as OpenClaw \citep{openclaw}, Nanobot \citep{nanobot}, and Claude Code \citep{claudecode} reflects a broader shift in large language models from single-turn question answering toward long-horizon assistants that support users across days, projects, and evolving context \citep{mialon2023gaia,li2026clawsbench}. 
In such settings, users rarely begin with a complete specification of what they actually need. Instead, they typically issue an \textbf{initial request}, a brief and often underspecified instruction that states only the surface goal, while the intended assistance also depends on complex and subtle \textbf{hidden intents} that users do not explicitly state, such as habits, constraints, and preferences. These intents can emerge gradually over long-horizon interactions, where an agent should integrate signals from multiple turns and reason over long information dependencies across sessions with the same user \citep{kim2026persona2web,lyu2026personalalign,liu2026perma}.

For instance, when a user asks ``\textit{help me plan a trip for next week}'' or ``\textit{prepare the client update deck}'', a strong assistant may use relevant information from a session three weeks earlier, such as travel preferences (e.g., budget, timing, and destinations) or deck conventions (e.g., format, metrics, and terminology), to proactively infer the user's hidden intents instead of waiting for specific instructions.
In practical applications, users expect agents to surface what needs clarification and decide what can be inferred, rather than treating underspecification as a reason to remain passive.
Addressing such requests requires \textbf{proactivity}: \textit{the ability to use goals, context, and prior interactions to anticipate user needs, recognize what remains underspecified, and move the task forward through appropriate action or clarification, while reducing the user’s operational and cognitive effort} \citep{sun2026ambibench,kong2026proactivemobile,kaur2026proper}. 
This capability shifts the assistant from passively following explicit instructions to actively managing underspecified tasks \citep{xie2026pask}.

However, proactive assistance in long-horizon personal assistant workflows remains underexplored. General agent benchmarks often assume explicit goals at interaction time \citep{liu2023agentbench,kim2026persona2web,zhang2026clawbench}. Memory benchmarks emphasize storing, retrieving, and applying prior information, while placing less focus on its role in uncovering and resolving underspecified requirements in long-horizon personal assistant workflows \citep{shen2026mem2actbench,liu2026memgui,he2026memoryarena}. Proactive benchmarks are mostly built around mobile or GUI settings with device context, visual trajectories, timely clarification, and short consumer tasks \citep{chen2026knowu,chai2026pira,nathani2026proactive}. In OpenClaw-style personal assistants, proactiveness takes a different form. Agents operate over persistent files and workspaces, coordinate tools to produce and revise artifacts, and maintain consistency with cross-session decisions and preferences. Missing requirements may surface only after intermediate deliverables are created, yet they can affect later file edits, artifact quality, and downstream task decisions \citep{vidgen2026apex,ji2026clawarena}.

To address this gap, we introduce \textbf{\bench}, a benchmark for evaluating proactive assistance in long-horizon personal assistant workflows. {\bench} places agents in persistent project environments where tasks unfold through multi-turn interaction, tool use, and iterative artifact creation. Each task begins with a natural but underspecified request, requiring the agent to identify hidden intents that capture user preferences and task dependencies. These intents may be revealed gradually through interaction, persist across sessions, and need to be reused in later tasks. For example, an assistant may need to apply a file format and naming convention established in a prior session to complete a later request without asking the user again. {\bench} captures this structure through \textit{100 multi-turn tasks across 5 domain-specific user personas, organized into multi-session workflows with cross-session dependencies}. We evaluate agents on both proactive assistance (Proactivity) and task completion (Completeness) by testing whether they address hidden intents early enough to support downstream decisions and complete the workflow successfully.

Our systematic experiments on nine frontier models reveal clear gaps in task completion and proactive intent resolution, distinguish task completeness from proactivity, and show substantial variation across domains and task types. Our main contributions:
\begin{itemize}[leftmargin=10pt, topsep=0pt, itemsep=1pt, partopsep=1pt, parsep=1pt]
    \item We formalize \emph{proactivity} for long-horizon personal agents.

    \item We introduce {\bench}, a benchmark for proactive assistance with 100 multi-turn tasks spanning five domain-specific personas, jointly evaluating proactivity and task completion via agent trajectories with long-range, cross-session dependencies.



    \item Extensive experiments show (1) proactive assistance remains challenging for frontier agents, (2) a clear distinction between completing tasks (completeness) and reducing user burden (proactivity), and (3) the value of prior interaction for proactive intent resolution in later tasks.
\end{itemize}



    

\section{Related Work}
\label{sec:related_work}

\paragraph{Personal Assistant Benchmarks.}
Personal assistant benchmarks evaluate end-to-end tool use in realistic web and computer environments \citep{mialon2023gaia,zhou2023webarena,drouin2024workarena}, with extensions to multimodal control and stateful planning \citep{xie2024osworld,malay2026enterpriseops}.
Recently, the rapid rise of OpenClaw \citep{openclaw} has pushed benchmarks toward long-horizon personal assistant workflows grounded in persistent workspaces and artifacts, spanning everyday online tasks and productivity settings \citep{zhang2026clawbench,li2026clawsbench}, including multi-day living-world coworkers \citep{clawmark}, with trustworthy evaluation \citep{ye2026claw} and robustness under evolving and conflicting information \citep{ji2026clawarena}.
Despite these advances, existing benchmarks rarely evaluate whether agents can proactively track, surface, and resolve hidden intents across multi-session workflows.

\paragraph{Memory Agent Benchmarks.}
Memory agent benchmarks evaluate whether agents can store, retrieve, and reuse user information across sessions \citep{shen2026mem2actbench,liu2026memgui,zhang2026memorycd}.
These benchmarks provide useful tests of long-term memory, personalization, and cross-session consistency \citep{he2026memoryarena,kim2026persona2web,liu2026perma}.
However, they usually treat memory as evidence for completing a known task, rather than as a signal for detecting missing requirements and deciding when to ask for clarification.
This leaves open how agents should use memory to detect underspecified requirements and resolve hidden intents as workflows evolve through interaction.
{\bench} addresses this gap with a broader evaluation setting that combines memory, workspace state, and interaction history to assess proactivity and task completeness in long-horizon personal assistant workflows.

\paragraph{Proactive Evaluation.}
Proactive benchmarks mainly study mobile or GUI agents, where proactivity is framed as using device context, interaction traces, and visual states to infer underspecified needs, ask clarifying questions, or intervene during app usage \citep{chen2026knowu,sun2026ambibench,nathani2026proactive}.
They often emphasize short-horizon everyday tasks with clear endpoints, such as booking and ordering \citep{kong2026proactivemobile,chai2026pira,nie2026pspa}. 
This leaves professional workflows and artifact-centered tasks underexplored, especially cases in which missing requirements may affect later edits or project decisions \citep{openclaw,zhang2026clawbench}.
In contrast, {\bench} focuses on long-horizon personal assistance in persistent workspaces where hidden intents may emerge late and earlier artifacts directly determine downstream decisions.

\section{Benchmark}
\label{sec:bench}

\begin{figure*}[tb!]
    \centering
    \includegraphics[width=1.0\linewidth]{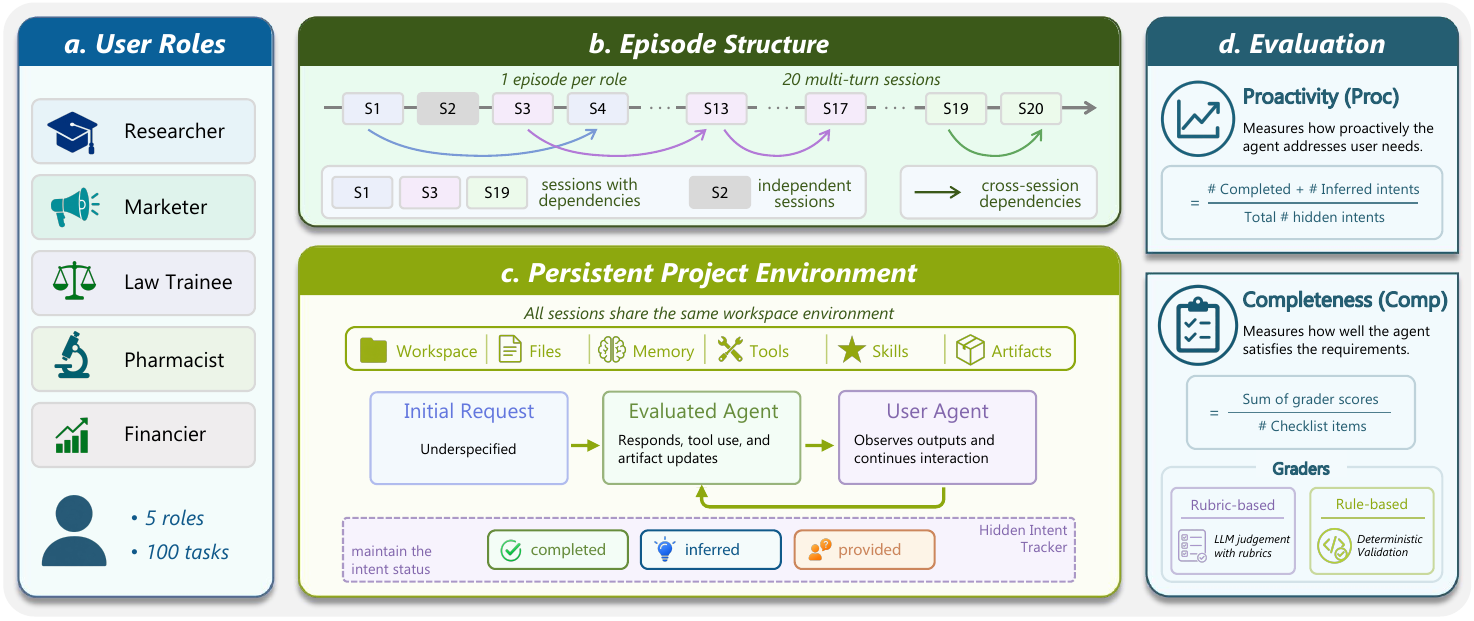}
    \vspace{-13 pt}
    \caption{Overview of {\bench}.}
    \label{fig:process}
    \vspace{-15 pt}
\end{figure*}

In this section, we present the design of {\bench}, as illustrated in Fig.~\ref{fig:process}. We target long-horizon personal assistant workflows in persistent project environments, where each session begins with a natural but underspecified request. Missing requirements may emerge after intermediate artifacts are produced and the interaction progresses, while preferences may carry across sessions and shape later decisions \citep{kim2026persona2web,lyu2026personalalign,liu2026perma}. {\bench} includes five user roles across distinct domains (researcher, marketer, law trainee, pharmacist, and financier), covering diverse workflows and constraints. For each role, we construct one episode with 20 sessions, where each session corresponds to one multi-turn task. We organize these tasks into multi-session episodes with cross-session dependencies, and evaluate agents on both proactive intent resolution (Proactivity) and task completion (Completeness).

\subsection{Evaluated Agent System}
\label{sec:evaluated_agent_system}

\paragraph{Agent paradigm.}
We focus on \emph{long-horizon personal agents} that assist users in both professional and everyday knowledge work by planning, producing, and refining concrete artifacts such as code, documents, and structured outputs \citep{openclaw,zhang2026clawbench,li2026clawsbench}. These agents typically adopt a modular design, where capabilities are composed from reusable components in a ReAct style \citep{yao2022react}, including tool interfaces, skills, and workspace operations. In our setting, the agent acts over a persistent project environment and makes progress mainly by iteratively updating intermediate artifacts.

\paragraph{Sessions.}
A user interacts with the agent through multiple {sessions}, and each session is a multi-turn conversation aimed at completing one task. Sessions share the same project workspace, so relevant files, intermediate artifacts, and prior outputs can carry over when appropriate. When needed, the agent may also consult memory to retain user-specific preferences or earlier decisions and apply them consistently in later sessions.

\paragraph{Tools and skills.}
Personal assistant agents operate in real-world environments where progress relies on invoking external tools and reusable skills to manipulate persistent artifacts \citep{schick2023toolformer,yao2024tau,patil2024gorilla,wang2026skillx}.
Accordingly, our tasks are grounded in practical tool and skill interfaces, such as shopping tool, web search tool, and data processing skill. This design requires the agent to coordinate tool calls and skill invocation to iteratively refine artifacts and produce task-ready outputs.

\subsection{User Agent}
\label{sec:user_agent}

\paragraph{User roles.}
{\bench} is centered on a {user agent} that simulates one user over an extended period. Each user is specified by a role that captures stable attributes, including occupation, routines, preferences, working style, and long-term goals \citep{kim2026persona2web}. 
Roles are constructed with domain experts to ensure realism and sufficient specificity, and are then lightly normalized to keep granularity and coverage consistent across users. \textit{Throughout the paper, we denote the evaluated system as the agent and the simulated counterpart as the user.}

\paragraph{Episodes.}
For each user agent, we define an \textbf{episode} that simulates the user’s long-horizon workflow across multiple tasks. Each episode contains 20 sessions, and each session corresponds to one task addressed through multi-turn interaction. Across sessions, the agent may leverage memory to carry forward relevant information when needed.

\paragraph{Workflow-grounded task design.}
To model OpenClaw-style assistance, {\bench} builds tasks around \emph{persistent workspace artifacts}, \emph{long-horizon professional workflows}, and \emph{recoverable hidden intents}.\footnote{A hidden intent is recoverable when it is absent from the initial request but can still be inferred or elicited from evidence available to the agent (e.g., prior sessions, workspace artifacts, or targeted clarification).} Each instance is derived from domain experts' authentic work routines and supporting materials, then shaped to require producing or revising concrete deliverables in the project environment. Progress depends on reading or updating files, repairing drafts, synthesizing evidence across documents, coordinating tools or skills, and preserving conventions from earlier sessions. Human experts further review each task to ensure it is realistic, feasible with the available files, tools, skills, and graders, and grounded in a correct and well-scoped workflow. These characteristics are reflected in App.~\ref{app:benchmark_statistics} and illustrated by case studies in App.~\ref{app:case_study}.

\paragraph{Dependency structure.}
In long-horizon use, sessions are not always independent, as later requests may rely on information from earlier interactions \citep{he2026memoryarena}. We therefore incorporate cross-session dependencies within each episode. Among the 20 tasks, we include (1) six strong dependency groups, each comprising two to three tasks that share essential carry-over information for successful completion, and (2) five largely independent tasks that broaden coverage of stand-alone workflows. In the latter case, any dependencies are lightweight and typically reflect general preferences, such as applying a consistent file naming convention or output directory structure.

\subsection{Task Formulation}
\label{sec:task_formulation}

\paragraph{Initial request.}
Users rarely begin a session with a complete specification of what they ultimately need. Instead, they usually provide a short, goal-oriented prompt and refine requirements as the agent produces intermediate artifacts and asks targeted questions \citep{vijayvargiya2026asking,chen2024learning,kaur2026proper}. Accordingly, each session in {\bench} starts with an \textbf{initial request} $u_1$ that initiates the task.
The initial request is designed to be natural and contextually plausible, while remaining minimally sufficient to enable progress and preserve realistic underspecification. In addition to \textit{user-issued messages}, we also allow \textit{environment-triggered signals} to start a session, such as external structured inputs or agent heartbeats \citep{nanobot} that the agent should recognize and respond to proactively.

\paragraph{Hidden intents.}
To formalize underspecification, each task is annotated with a set of {hidden intents} $\mathcal{I}=\{i_1,\dots,i_m\}$.
Each intent $i$ represents a latent requirement that should shape how the task is handled, e.g., constraints, preferences, and downstream dependencies. Hidden intents can be session-local or persistent across sessions. The agent can satisfy an intent by inferring it from prior interaction and memory, or by asking a focused question that elicits the missing requirement and then acting on it.

\paragraph{Checklist.}
For each task, we provide a checklist $\mathcal{C}=\{c_1,\dots,c_n\}$ that defines verifiable completion criteria for the final outcome and required artifacts. Checklist items specify what should be delivered, including files to create or modify, fields to populate, outputs to generate, and constraints to satisfy. During data construction, human experts invest substantial effort to execute and review each task, produce reference solutions, and ensure that checklist items are both necessary and sufficient.
Compared with hidden intents, which capture latent preferences or constraints, checklist items are more concrete and fine-grained, often with ground-truth-like verification logic that defines explicit obligations the agent must fulfill. A more detailed distinction between hidden intents and checklists is provided in App.~\ref{app:hidden_intents_checklists}.

\paragraph{Graders.}
We implement checklist verification with two types of graders:
\begin{itemize}[leftmargin=10pt, topsep=0pt, itemsep=1pt, partopsep=1pt, parsep=1pt]
    \item \textbf{Rubric-based evaluation}. For open-ended content where deterministic checks are unsuitable, we use rubric-based model evaluation to assess whether the output satisfies task requirements and user constraints.
    \item \textbf{Rule-based verification}. For objective conditions, we apply deterministic rule-based verification, such as file existence, exact string matching, correct tool use, and schema validity.
\end{itemize}

\subsection{Session Interaction and Intent Tracking}
\label{sec:interaction_tracking}

\begin{figure*}[tb!]
    \centering
    \includegraphics[width=1.0\linewidth]{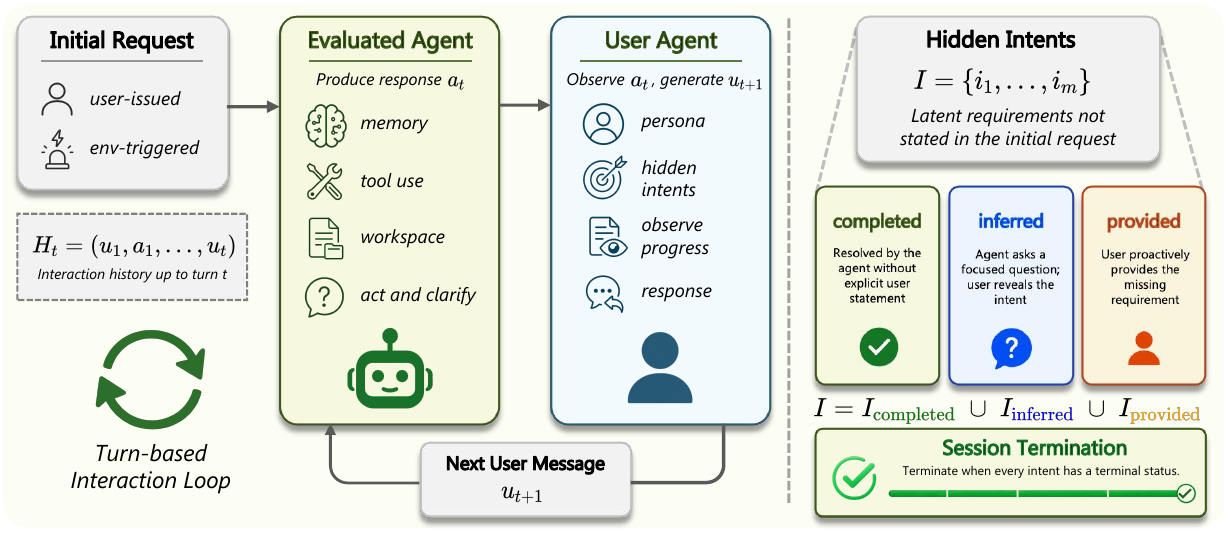}
    \vspace{-12 pt}
    \caption{Overview of one benchmark session.
The evaluated agent interacts with a simulated user agent in a turn-based loop, while the user agent tracks hidden intents and assigns each intent a terminal status as {\completed}, {\inferred}, or {\provided}.
The session ends when all hidden intents have reached terminal status.}
    \label{fig:interaction}
    \vspace{-16 pt}
\end{figure*}

\paragraph{Interaction process.}
Fig.~\ref{fig:interaction} illustrates the turn-based loop for one benchmark session with hidden intents $\mathcal{I}=\{i_1,\dots,i_m\}$ and checklist $\mathcal{C}
  =\{c_1,\dots,c_n\}$.
Each session starts with an \textit{initial request} $u_1$ that initiates the task. At each turn, the agent produces a response $a_t$, which may involve tool use and artifact creation or updates in the workspace. The user agent then observes the agent response together with any newly produced or updated artifacts, updates the tracking state for hidden intents in $\mathcal{I}$ by assigning terminal statuses when applicable, and generates the next user message. If the agent asks a relevant question, the user agent answers it. Otherwise, if some requirements remain underspecified, the user agent proactively provides the missing task-relevant information to keep the task moving. The interaction proceeds in this alternating manner until the session terminates.
Formally, let $u_t$ and $a_t$ denote the user and agent utterances at turn $t$. We write the interaction history up to turn $t$ as
\begin{equation}
H_t = (u_1, a_1, \dots, u_t),
\end{equation}
and use $H$ to denote the resulting session trajectory. At each turn, the agent produces a response $a_t$, including any tool calls and workspace updates. A session terminates when each intent in $\mathcal{I}$ has been assigned a terminal status and the agent has produced its final response. The full assignment procedure, response mechanism, and prompt template are provided in App.~\ref{app:user_agent}.


\paragraph{Intent status assignment.}
Each hidden intent $i\in\mathcal{I}$ is initially unstated in the initial request. As the interaction unfolds along $H$, we assign $i$ exactly one terminal status from the set $\{\completed,\inferred,\provided\}$:
\begin{itemize}[leftmargin=10pt, topsep=0pt, itemsep=1pt, partopsep=1pt, parsep=1pt]
    \item $\completed$: the agent resolves $i$ without the user explicitly stating it, by producing an action or artifact consistent with the intent.
    \item $\inferred$: the agent asks a focused question that directly targets $i$, and the user reveals the missing requirement in the next turn, after which the agent can act on it.
    \item $\provided$: the agent neither resolves $i$ nor asks a relevant question, and the user must proactively supply $i$ to move the task forward.
\end{itemize}

\paragraph{Session termination.}
Once an intent is assigned a terminal status, it is excluded from further tracking within the same session. 
A session terminates only when every intent in $\mathcal{I}$ has been assigned a terminal status and the agent has produced its final response. 
At this point, each hidden intent has either been completed by the agent, elicited through a clarification, or provided by the user. The user agent has no further hidden information to provide, and the interaction has reached a natural stopping point.
Let {\intentcompleted}, {\intentinferred}, and {\intentprovided} denote the subsets of $\mathcal{I}$ assigned to $\completed$, $\inferred$, and $\provided$ under $H$, respectively, so that
\begin{equation}
\mathcal{I}={\intentcompleted}\cup {\intentinferred}\cup {\intentprovided}
\quad\text{and}\quad
{\intentcompleted},{\intentinferred},{\intentprovided}\ \text{are disjoint.}
\end{equation}

\subsection{Evaluation Protocol}
\label{sec:evaluation}

We evaluate each agent on both \textbf{proactivity} and \textbf{completeness}, which measure whether the agent resolves hidden intents proactively and ultimately satisfies the task's verifiable requirements. Detailed evaluation protocols are provided in App.~\ref{app:detailed_evaluation}.


\paragraph{Proactivity.}
We define the proactivity score as the fraction of intents that the agent resolves proactively, either by direct completion or by targeted elicitation,
\begin{equation}
\textsc{Proc}(H)=\frac{\left|{\intentcompleted}\right|+\left|{\intentinferred}\right|}{|\mathcal{I}|}.
\end{equation}
The score is designed to separate agent-driven requirement discovery ($\intentinferred$ and $\intentcompleted$) from user-driven disclosure ($\intentprovided$). It captures whether the agent goes beyond the surface request to identify what remains underspecified and reduce the user’s operational and cognitive effort through appropriate action or clarification. We give $\intentcompleted$ and $\intentinferred$ equal credit because both reflect agent initiative: some intents can be addressed directly, while others should be resolved through targeted clarification.

\paragraph{Completeness.}
Completeness measures whether the agent ultimately satisfies the task’s verifiable requirements over the course of a session. For each checklist item $c\in\mathcal{C}$, we compute a grader score $s(c,H)\in\{0,1\}$ using either a deterministic program or rubric-based model evaluation, following Sec.~\ref{sec:task_formulation}. Using $H$ allows the grader to incorporate evidence accumulated across turns, including intermediate artifacts and partial progress produced at different points in the interaction. We then define the task completeness score as
\begin{equation}
\textsc{Comp}(H)=\frac{1}{|\mathcal{C}|}\sum_{c\in\mathcal{C}} s(c,H).
\end{equation}
\paragraph{Metric relationship.}
Proactivity and completeness capture related but distinct aspects of agent behavior. In our protocol, the simulated user eventually provides any hidden intent that the agent fails to elicit or address, so the final trajectory can contain the full set of intents even when the agent passively waits for user-provided information. Thus, \textsc{Proc} measures how much the agent drives requirement discovery, while \textsc{Comp} measures whether the agent turns the resulting trajectory into correct artifacts and decisions. The two scores can therefore diverge substantially and reflect different capabilities. This separation is analyzed in Sec.~\ref{sec:analysis} and discussed further in App.~\ref{app:metric_relationship}.

\section{Experiments}
\label{sec:exp}

\subsection{Setup}
\label{sec:exp_setup}

\paragraph{Model Setup.}
We evaluate nine frontier LLMs spanning distinct model families: GPT-5.4 \citep{gpt54}, Gemini-3.1 Pro \citep{gemini31pro}, Claude 4.6 Opus \citep{claudeopus46}, DeepSeek V3.2 \citep{liu2025deepseek}, MiniMax M2.7 \citep{minimax2026m27}, Kimi K2.5 \citep{team2026kimi}, Seed2.0 Pro \citep{seed20pro}, GLM-5.1 \citep{zai2026glm51}, and Qwen3.6 Plus \citep{qwen36plus}. 
All models are evaluated under the same agentic scaffold, adapted from Nanobot \citep{nanobot}, so that performance differences primarily reflect model capability rather than scaffold-specific components.

\paragraph{Environment Setup.}
We use the default decoding parameters for all models and enable thinking. We report $\textsc{Proc}$ and $\textsc{Comp}$ as the main metrics. Since long-horizon agent behavior can be stochastic \citep{mustahsan2025stochasticity}, we run each task three times with independent trajectories and report averaged results with standard deviations for robust estimation. We use GPT-5.4 as the base model for the user agent and as the rubric-based grader, with temperature set to zero.

\subsection{Main Results}
\label{sec:exp_main}

{
\definecolor{OverallAvgComp}{HTML}{FFF4CC}
\definecolor{OverallAvgProc}{HTML}{D9F0FF}

\newcommand{\overallavg}[2]{#1$_{\scriptscriptstyle #2}$}
\newcommand{\overallpair}[2]{#1 / #2}

\begin{table*}[tb]
\centering
\caption{Overall results for \textsc{Proc} / \textsc{Comp} (\%). Results are averaged over three runs, with subscripts denoting standard deviations.}
\vspace{-6 pt}
\label{tab:overall_results}
\scriptsize
\resizebox{\linewidth}{!}{
\begin{tabular}{lcc|ccccc}
\toprule[1.25pt]
\multirow{2}{*}{\textbf{Model}} &
\multicolumn{2}{c|}{\textbf{Average}} &
\multicolumn{5}{c}{\textbf{User domain}} \\
\cmidrule(lr){2-3} \cmidrule(lr){4-8}
& \cellcolor{OverallAvgProc}{\textbf{\textsc{Proc}}} &
\cellcolor{OverallAvgComp}{\textbf{\textsc{Comp}}} &
\textbf{Researcher} &
\textbf{Marketer} &
\textbf{Pharmacist} &
\textbf{Law Trainee} &
\textbf{Financier} \\
\midrule[0.75pt]
GPT-5.4 &
\cellcolor{OverallAvgProc}{\overallavg{\textbf{67.0}}{2.1}} &
\cellcolor{OverallAvgComp}{\overallavg{65.6}{1.8}} &
\overallpair{46.0}{66.4} &
\overallpair{\textbf{78.2}}{67.1} &
\overallpair{75.9}{71.5} &
\overallpair{\textbf{56.9}}{\textbf{61.9}} &
\overallpair{\textbf{78.1}}{61.2} \\
Gemini 3.1 Pro &
\cellcolor{OverallAvgProc}{\overallavg{57.1}{0.9}} &
\cellcolor{OverallAvgComp}{\overallavg{60.0}{0.8}} &
\overallpair{41.1}{59.2} &
\overallpair{65.0}{62.1} &
\overallpair{71.0}{72.1} &
\overallpair{50.0}{55.3} &
\overallpair{58.6}{51.1} \\
Claude Opus 4.6 &
\cellcolor{OverallAvgProc}{\overallavg{65.5}{\mathrm{1.4}}} &
\cellcolor{OverallAvgComp}{\overallavg{\textbf{67.6}}{\mathrm{1.5}}} &
\overallpair{\textbf{50.3}}{\textbf{74.5}} &
\overallpair{75.0}{\textbf{74.6}} &
\overallpair{\textbf{82.8}}{68.6} &
\overallpair{45.7}{57.2} &
\overallpair{73.8}{\textbf{63.2}} \\
DeepSeek V3.2 &
\cellcolor{OverallAvgProc}{\overallavg{53.3}{1.9}} &
\cellcolor{OverallAvgComp}{\overallavg{57.8}{3.0}} &
\overallpair{29.0}{66.9} &
\overallpair{69.1}{59.4} &
\overallpair{75.9}{62.6} &
\overallpair{33.2}{51.1} &
\overallpair{59.1}{48.9} \\
MiniMax M2.7 &
\cellcolor{OverallAvgProc}{\overallavg{55.6}{3.2}} &
\cellcolor{OverallAvgComp}{\overallavg{60.0}{1.8}} &
\overallpair{33.4}{63.9} &
\overallpair{71.9}{61.9} &
\overallpair{77.1}{63.6} &
\overallpair{38.6}{52.5} &
\overallpair{57.2}{58.1} \\
Kimi K2.5 &
\cellcolor{OverallAvgProc}{\overallavg{43.1}{0.2}} &
\cellcolor{OverallAvgComp}{\overallavg{61.6}{1.9}} &
\overallpair{28.9}{63.5} &
\overallpair{41.2}{62.3} &
\overallpair{70.1}{\textbf{74.8}} &
\overallpair{34.8}{54.4} &
\overallpair{40.4}{52.9} \\
Seed2.0 Pro &
\cellcolor{OverallAvgProc}{\overallavg{58.4}{0.9}} &
\cellcolor{OverallAvgComp}{\overallavg{52.1}{3.8}} &
\overallpair{38.9}{59.6} &
\overallpair{71.4}{44.2} &
\overallpair{77.0}{67.6} &
\overallpair{46.0}{44.7} &
\overallpair{58.7}{44.5} \\
GLM-5.1 &
\cellcolor{OverallAvgProc}{\overallavg{58.4}{0.8}} &
\cellcolor{OverallAvgComp}{\overallavg{63.6}{2.9}} &
\overallpair{41.8}{61.6} &
\overallpair{62.6}{69.1} &
\overallpair{75.2}{70.3} &
\overallpair{45.5}{57.3} &
\overallpair{66.7}{59.8} \\
Qwen3.6 Plus &
\cellcolor{OverallAvgProc}{\overallavg{64.0}{1.1}} &
\cellcolor{OverallAvgComp}{\overallavg{64.1}{0.6}} &
\overallpair{40.1}{70.0} &
\overallpair{77.5}{66.6} &
\overallpair{79.7}{70.2} &
\overallpair{45.7}{60.2} &
\overallpair{77.1}{53.6} \\
\bottomrule[1.25pt]
\end{tabular}
}
\vspace{-10pt}
\end{table*}
}

Table~\ref{tab:overall_results} reports the aggregate performance of all evaluated models across the five user domains.

\paragraph{Model trends.}
The aggregate results show that {\bench} is challenging and discriminative. Across the nine models, average \textsc{Comp} ranges from 52.1 to 67.6, while average \textsc{Proc} ranges from 43.1 to 67.0, leaving room for further progress. GPT-5.4 achieves the highest average \textsc{Proc} at 67.0, while Claude 4.6 Opus obtains the highest average \textsc{Comp} at 67.6. Qwen3.6 Plus is also competitive on both dimensions, with 64.0 \textsc{Proc} and 64.1 \textsc{Comp}. The standard deviations are generally small, with most values below 2.0, which indicates that the aggregate trends are stable across repeated runs. Notably, Qwen3.6 Plus maintains low standard deviations while achieving strong scores on both metrics. At the same time, the rankings are not identical across metrics. For example, Kimi K2.5 reaches 61.6 \textsc{Comp} but only 43.1 \textsc{Proc}, whereas Seed2.0 Pro reaches 58.4 \textsc{Proc} but 52.1 \textsc{Comp}. This suggests that completing visible workflow requirements and proactively handling underspecified user needs remain related but distinct capabilities.

\paragraph{Domain trends.}
Performance varies substantially across user domains. Pharmacist tasks are the easiest overall, with the highest average \textsc{Proc} and \textsc{Comp}. This is consistent with many Pharmacist instances in {\bench} being grounded in concrete local files, literature summaries, lab records, and domain-specific skills needed for the workflow. Researcher tasks show a different pattern, with relatively low \textsc{Proc} despite strong \textsc{Comp}. This domain often involves research planning, rebuttal preparation, literature synthesis, and other challenging tasks with less standardized workflows. Law Trainee and Financier have the lowest average \textsc{Comp}. Their workflows often require risk-oriented legal or financial judgment, which makes task completion more difficult. These results show that {\bench} captures variation across domains. Concrete case studies are provided in App.~\ref{app:case_study}.

\subsection{Analysis}
\label{sec:analysis}

\begin{figure}[tb]
    \centering
    \includegraphics[width=0.98\linewidth]{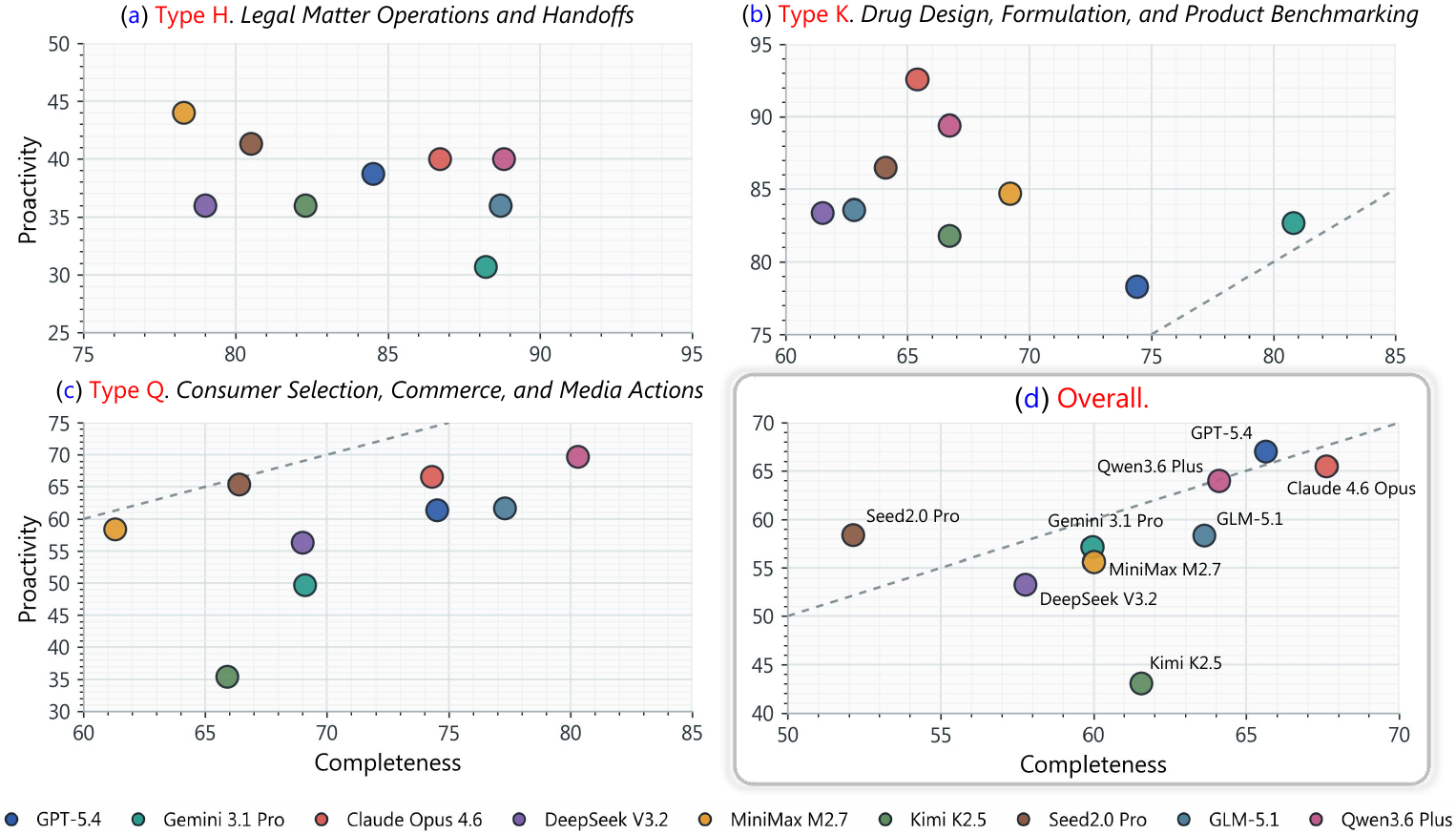}
    \caption{\textsc{Comp} and \textsc{Proc} for ({\color{blue}a, b, c}) three representative task workflow categories following the fine-grained taxonomy in Tab.~\ref{tab:task_taxonomy} of App.~\ref{app:taxonomy_of_benchmark_tasks}, and ({\color{blue}d}) the overall average across all tasks. The gray dashed line indicates \textsc{Comp} = \textsc{Proc}.}
    \label{fig:task_type_scatter}
    \vspace{-14 pt}
\end{figure}

\paragraph{Performance by task type.}
Fig.~\ref{fig:task_type_scatter} ({a, b, c}) compares \textsc{Comp} and \textsc{Proc} on three representative task categories. Detailed per-category results are reported in App.~\ref{app:exp_task_type} and Tab.~\ref{tab:task_type_scores}. Legal matter operations and handoffs (H) show the largest gap, with high average \textsc{Comp} but low average \textsc{Proc} (84.1\% vs. 38.1\%). In these cases, agents can often draft the requested document, but they fail to ensure that the matter is ready for handoff, leaving hidden intents about missing materials, blockers, and follow-up actions for the user to surface. Similarly, consumer selection, commerce, and media actions (Q) are more completion-oriented (70.8\% \textsc{Comp} vs. 58.2\% \textsc{Proc}). Drug design, formulation, and product benchmarking (K) show the opposite pattern, with higher \textsc{Proc} than \textsc{Comp} (84.9\% vs. 68.0\%). Here, hidden intents are often grounded in concrete scientific constraints, such as assumptions and experimental evidence, which agents can infer more easily than they can produce a fully comprehensive technical synthesis. Overall, performance varies across task categories and metrics, suggesting that different workflow structures stress different model capabilities.

\paragraph{Distinguishing proactivity from completeness.}
Fig.~\ref{fig:task_type_scatter} ({d}) compares aggregate \textsc{Proc} and \textsc{Comp} from Tab.~\ref{tab:overall_results}. The two metrics are positively related, as several high-scoring models appear near the upper-right region, but they measure different aspects of assistance. Since {\bench} continues until each hidden intent is either resolved by the agent or supplied by the simulated user, a reactive model can still recover a reasonable \textsc{Comp} score after the missing requirements become explicit. \textsc{Proc} instead measures whether the agent reduces this burden by resolving hidden intents or eliciting them before the user has to provide them.

The off-diagonal cases make this distinction clear. Kimi K2.5 attains a relatively high \textsc{Comp} score but a much lower \textsc{Proc} score, suggesting that it can execute tasks once constraints are stated, yet often waits for the user to reveal those constraints step by step. \textbf{This contrast reveals a practical decoupling between proactivity and completeness, where reactive recovery can preserve final task quality while shifting requirement discovery back to the user.} Seed2.0 Pro shows the reverse pattern, with higher \textsc{Proc} than \textsc{Comp}, indicating that early discovery of hidden intents is not sufficient when final execution remains weak. 
This behavior is consistent with our evaluation design, where hidden intents capture latent interaction requirements and checklist items capture verifiable obligations in the final outcome (App.~\ref{app:hidden_intents_checklists}). By making this gap observable, {\bench} tests whether agents merely complete underspecified workflows after user intervention or proactively reduce the user burden while moving those workflows toward successful completion.

\begin{wrapfigure}{r}{0.55\textwidth}
    \centering
    \includegraphics[width=\linewidth]{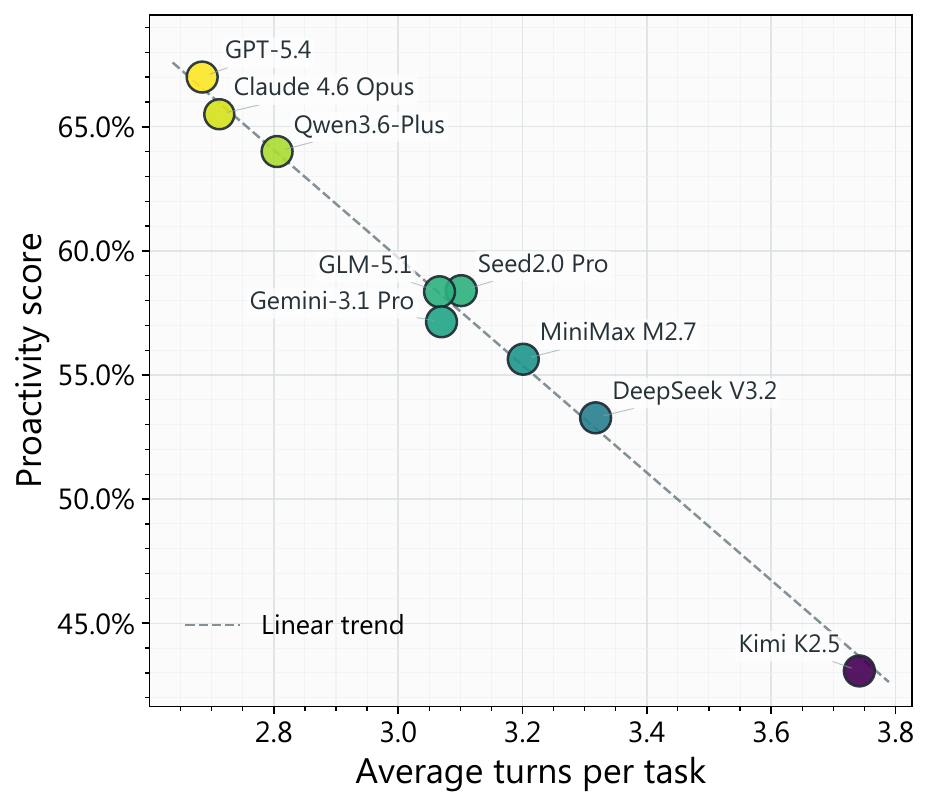}
    \vspace{-16 pt}
    \caption{Relationship between average interaction turns per task and model-level \textsc{Proc}. Each point denotes one model.}
    \label{fig:turn_count}
    \vspace{-8 pt}
\end{wrapfigure}
\paragraph{Turn count and interaction cost.}
Turn count provides an observable measure of the interaction cost. In {\bench}, a session continues until each hidden intent receives a terminal status ({\intentcompleted}, {\intentinferred}, and {\intentprovided}). More turns therefore indicate that the agent needed additional user input before the task became fully specified. This cost is not identical to user burden, since some extra turns may come from useful clarifications that reduce ambiguity. We therefore treat turn count as \textit{a complementary measure rather than a substitute for \textsc{Proc}}. 
Fig.~\ref{fig:turn_count} shows a negative association. GPT-5.4, Claude 4.6 Opus, and Qwen3.6 Plus lie in the low-turn and high-\textsc{Proc} region, suggesting that they more often resolve hidden intents through early action or focused elicitation. Kimi K2.5 has the highest average turn count and the lowest \textsc{Proc}, which indicates that its trajectories more often depend on user-supplied information. This pattern is consistent with the results in Sec.~\ref{sec:exp_main} and supports the design goal of {\bench}. A proactive assistant should improve final task outcomes while minimizing avoidable interaction needed to uncover and address the user's unstated requirements.

\begin{wrapfigure}{r}{0.47\textwidth}
    \vspace{-12 pt}
    \centering
    \includegraphics[width=\linewidth]{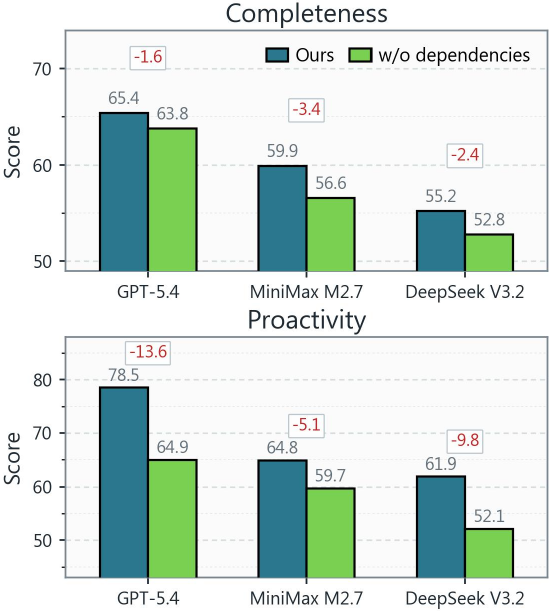}
    \vspace{-12 pt}
    \caption{Ablation on the final task of each strong dependency group: six per user, aggregated across five roles. \emph{Ours} uses the original trajectories, while \emph{w/o dependencies} removes preceding sessions from the same group.}
    \label{fig:ablation_study}
    \vspace{-16 pt}
\end{wrapfigure}
\paragraph{Prior interactions support proactive intent resolution.}
We ablate each strong dependency group to test whether earlier sessions help agents resolve later hidden intents. For each group, we remove the preceding sessions and evaluate only the final task. As shown in Fig.~\ref{fig:ablation_study}, removing history substantially reduces \textsc{Proc} while leaving \textsc{Comp} mostly stable. GPT-5.4 drops from 78.5 to 64.9, MiniMax M2.7 from 64.8 to 59.7, and DeepSeek V3.2 from 61.9 to 52.1 on \textsc{Proc}, with an average decrease of 9.5 points. By contrast, \textsc{Comp} decreases by only 2.5 points on average. This suggests that the preceding sessions are useful for resolving hidden intents before the user spells them out. Once those sessions are removed, agents can still recover some final-task quality through later user feedback, but they lose much of the ability to act proactively. This ablation confirms the importance of prior interaction for proactive intent resolution in later tasks.

\paragraph{More experiments.}
We provide additional analyses in App.~\ref{app:experiments}.
The judge reliability study (App.~\ref{app:judge_reliability}) shows low disagreement (< 4\%) under expert audits and independent frontier model audits.
App.~\ref{app:hidden_intents_status} reports hidden-intent terminal status distributions, App.~\ref{app:exp_task_type} breaks down performance by task type, and App.~\ref{app:failure_analysis} summarizes failure patterns with concrete examples from App.~\ref{app:case_study}.

\section{Conclusions}
\label{sec:conclusions}

We introduced {\bench}, a benchmark for evaluating proactive personal assistant agents in long-horizon workflows, covering 100 multi-turn tasks across five domain-specific personas with hidden intents, inter-task dependencies, and cross-session continuity. By jointly evaluating proactivity and completeness, {\bench} tests whether agents can resolve underspecified requests, reuse prior context, and produce task-ready artifacts in persistent workspaces.
Experiments on nine frontier models reveal clear gaps, distinguish proactivity from completeness, and show that prior interaction history helps agents resolve hidden intents while reducing avoidable user interaction.

\section{Limitations}
\label{sec:limitations}

Our benchmark has several inherent limitations. The users are simulated rather than real humans, which is difficult to avoid because long horizon evaluation with live users is costly, hard to reproduce, and difficult to scale.
In addition, our experiments use a single agentic scaffold adapted from Nanobot, which provides a controlled evaluation setup but may not capture the full variation introduced by alternative scaffolds, which can add substantial adaptation effort and scaffold-specific confounds.

{\small
\bibliographystyle{plainnat}
\bibliography{reference}
}


\clearpage
\appendix

\part*{Appendix}
\addcontentsline{toc}{part}{Appendix}  

\startcontents[appendix]
\printcontents[appendix]{}{1}{}

\newpage

\section{Benchmark Statistics}
\label{app:benchmark_statistics}

\subsection{Task and Grader Statistics}
\label{app:user_level_task_grader_statistics}

In this subsection, we summarize the task and grader statistics for the 100 tasks across all users, as shown in Tab.~\ref{tab:benchmark_user_statistics}.
\begin{table}[htbp]
\centering
\small
\setlength{\tabcolsep}{5pt}
\caption{Task and Grader Statistics.}
\label{tab:benchmark_user_statistics}
\begin{tabular}{cccccc}
\toprule
\textbf{User} & \textbf{Task} & \textbf{Session} & \textbf{Hidden Intent} & \textbf{Rubric} & \textbf{\shortstack{Rule}} \\
\midrule
Researcher & 20 & 20 & 84 & 125 & 31 \\
Marketer & 20 & 20 & 74 & 58 & 11 \\
Law Trainee & 20 & 20 & 89 & 131 & 43 \\
Pharmacist & 20 & 20 & 195 & 85 & 25 \\
Financier & 20 & 20 & 82 & 111 & 58 \\
\midrule
\textbf{Total} & \textbf{100} & \textbf{100} & \textbf{524} & \textbf{510} & \textbf{168} \\
\bottomrule
\end{tabular}
\end{table}

\subsection{Tool and Skill Statistics}
\label{app:tool_skill_inventory}

Tabs.~\ref{tab:benchmark_tools} and \ref{tab:benchmark_skills} summarize the tool and skill inventory of the benchmark. In total, the benchmark contains 187 unique tools and 21 skills. 

\par\medskip
\begingroup
\footnotesize
\setlength{\LTpre}{8pt}
\setlength{\LTpost}{0pt}
\setlength{\tabcolsep}{3pt}
\begin{longtable}{@{}p{0.15\linewidth}p{0.27\linewidth}p{0.50\linewidth}@{}}
\caption{Benchmark tools grouped by application domain.}
\label{tab:benchmark_tools}\\
\toprule
\textbf{Domain} & \textbf{Tool} & \textbf{Function} \\
\midrule
\endfirsthead

\toprule
\textbf{Domain} & \textbf{Tool} & \textbf{Function} \\
\midrule
\endhead

\endfoot

\bottomrule
\endlastfoot

Amazon & Add Address & Add a saved address in Amazon. \\
 & Add Product to Cart & Add a product to the cart in Amazon. \\
 & Add Product to Wish List & Add a product to the wish list in Amazon. \\
 & Add to History & Add a product to browsing history in Amazon. \\
 & Clear Cart & Remove all items from the shopping cart in Amazon. \\
 & Log In & Authenticate in Amazon. \\
 & Log Out & Sign out of Amazon. \\
 & Move Cart Item to Wish List & Move a cart item to the wish list in Amazon. \\
 & Move Wish List Item to Cart & Move a wish-list item to the cart in Amazon. \\
 & Place Order & Complete a purchase in Amazon. \\
 & Post Product Question & Post a product question in Amazon. \\
 & Search Product Types & Search product categories in Amazon. \\
 & Search Products & Search products in Amazon. \\
 & Search Sellers & Search sellers in Amazon. \\
 & Update Cart Quantity & Change product quantity in the cart in Amazon. \\
 & View Account & View account information in Amazon. \\
 & View Addresses & View saved addresses in Amazon. \\
 & View Cart & View the shopping cart in Amazon. \\
 & View Last Purchase & View the most recent product purchase in Amazon. \\
 & View Order & View an order in Amazon. \\
 & View Orders & View orders in Amazon. \\
 & View Payment Card & View a stored payment card in Amazon. \\
 & View Payment Cards & View stored payment cards in Amazon. \\
 & View Product & View product in Amazon. \\
 & View Product Options & View available options for a product in Amazon. \\
 & View Product Purchases & View past purchases of a product in Amazon. \\
 & View Product Q\&A & View product questions and answers in Amazon. \\
 & View Product Questions & View product questions in Amazon. \\
 & View Product Reviews & View product reviews in Amazon. \\
 & View Recommendations & View recommended products in Amazon. \\
 & View Seller & View seller information in Amazon. \\
 & View Wish List & View the wish list in Amazon. \\
 & Write Product Review & Write a product review in Amazon. \\
\midrule
Gmail & Archive Thread & Archive an email thread in Gmail. \\
 & Create Draft & Create an email draft in Gmail. \\
 & Forward Thread & Forward an email thread in Gmail. \\
 & Label Thread & Apply a label to an email thread in Gmail. \\
 & Log In & Authenticate in Gmail. \\
 & Log Out & Sign out of Gmail. \\
 & Mark Read & Mark an email thread as read in Gmail. \\
 & Mark Unread & Mark an email thread as unread in Gmail. \\
 & Remove Label & Remove a label from an email thread in Gmail. \\
 & Reply to Email & Reply to an email in Gmail. \\
 & Search Labels & Search labels in Gmail. \\
 & Search Users & Search users in Gmail. \\
 & Send Draft & Send a previously saved draft in Gmail. \\
 & Send Email & Send an email in Gmail. \\
 & Sign Up & Create an account in Gmail. \\
 & Star Thread & Star an email thread in Gmail. \\
 & Update Draft & Edit an email draft in Gmail. \\
 & View Category Sizes & View category sizes in Gmail. \\
 & View Draft & View draft in Gmail. \\
 & View Drafts & View drafts in Gmail. \\
 & View Email & View email in Gmail. \\
 & View Inbox Threads & View inbox threads in Gmail. \\
 & View Outbox Threads & View outbox threads in Gmail. \\
 & View Starred Threads & View starred threads in Gmail. \\
 & View Thread & View thread in Gmail. \\
\midrule
Phone & Add Contact & Add contact in Phone. \\
 & Create Alarm & Create an alarm in Phone. \\
 & Get Date and Time & Retrieve the current date and time from Phone. \\
 & Log In & Authenticate in Phone. \\
 & Search Contacts & Search contacts in Phone. \\
 & Search Messages & Search text messages in Phone. \\
 & Send Text Message & Send a text message in Phone. \\
 & Update Alarm & Update an alarm in Phone. \\
 & Update Contact & Update contact in Phone. \\
 & View Alarm & View an alarm in Phone. \\
 & View Message Thread & View a text-message thread in Phone. \\
 & View Text Message & View a text message in Phone. \\
\midrule
Simple Note & Add Content to Note & Add content to note in Simple Note. \\
 & Create Note & Create note in Simple Note. \\
 & Delete Note & Delete note in Simple Note. \\
 & Log In & Authenticate in Simple Note. \\
 & Log Out & Sign out of Simple Note. \\
 & Search Notes & Search notes in Simple Note. \\
 & Update Note & Update note in Simple Note. \\
 & View Account & View account information in Simple Note. \\
 & View Note & View note in Simple Note. \\
 & View Profile & View profile in Simple Note. \\
\midrule
Splitwise & Add Group Member & Add a member to a group in Splitwise. \\
 & Create Group & Create group in Splitwise. \\
 & Delete Expense & Delete expense in Splitwise. \\
 & Log In & Authenticate in Splitwise. \\
 & Record Expense & Record an expense in Splitwise. \\
 & Search Users & Search users in Splitwise. \\
 & Update Expense & Update expense in Splitwise. \\
 & View Group & View group in Splitwise. \\
 & View Group Balance & View the balance of a group in Splitwise. \\
 & View Group Expenses & View expenses recorded in a group in Splitwise. \\
 & View Groups & View groups in Splitwise. \\
 & View Member Balances & View per-member balances in Splitwise. \\
 & View Person Balance & View the balance of one person in Splitwise. \\
\midrule
Spotify & Add Song to Playlist & Add a song to a playlist in Spotify. \\
 & Add to Queue & Add an item to the playback queue in Spotify. \\
 & Create Playlist & Create playlist in Spotify. \\
 & Follow Artist & Follow an artist in Spotify. \\
 & Log In & Authenticate in Spotify. \\
 & Log Out & Sign out of Spotify. \\
 & Play Music & Start music playback in Spotify. \\
 & Previous Song & Return to the previous song in Spotify. \\
 & Remove Downloaded Song & Remove a downloaded song in Spotify. \\
 & Remove Song from Playlist & Remove a song from a playlist in Spotify. \\
 & Remove Song from Queue & Remove a song from the playback queue in Spotify. \\
 & Review Song & Write a song review in Spotify. \\
 & Save Album & Save an album in Spotify. \\
 & Save Album to Library & Save an album to the library in Spotify. \\
 & Save Playlist & Save a playlist in Spotify. \\
 & Save Song & Save a song in Spotify. \\
 & Save Song to Library & Save a song to the library in Spotify. \\
 & Search Albums & Search albums in Spotify. \\
 & Search Artists & Search artists in Spotify. \\
 & Search Playlists & Search playlists in Spotify. \\
 & Search Songs & Search songs in Spotify. \\
 & Search Users & Search users in Spotify. \\
 & Update Song Review & Edit a song review in Spotify. \\
 & View Album & View album in Spotify. \\
 & View Album Library & View album library in Spotify. \\
 & View Artist & View artist in Spotify. \\
 & View Current Song & View the currently playing song in Spotify. \\
 & View Followed Artists & View artist following in Spotify. \\
 & View Genres & View genres in Spotify. \\
 & View Liked Songs & View liked songs in Spotify. \\
 & View Playlist & View playlist in Spotify. \\
 & View Playlist Library & View playlist library in Spotify. \\
 & View Queue & View the playback queue in Spotify. \\
 & View Song & View song in Spotify. \\
 & View Song Library & View song library in Spotify. \\
 & View Song Review & View a song review in Spotify. \\
 & View Song Reviews & View song reviews in Spotify. \\
\midrule
Todoist & Add Task Label & Add a label to a task in Todoist. \\
 & Assign Task & Assign or unassign a task in Todoist. \\
 & Create Label & Create label in Todoist. \\
 & Create Project & Create project in Todoist. \\
 & Create Section & Create section in Todoist. \\
 & Create Subtask & Create a subtask in Todoist. \\
 & Create Task & Create task in Todoist. \\
 & Delete Label & Delete label in Todoist. \\
 & Delete Notifications & Delete notifications in Todoist. \\
 & Delete Project & Delete project in Todoist. \\
 & Delete Section & Delete section in Todoist. \\
 & Delete Subtask & Delete a subtask in Todoist. \\
 & Delete Task & Delete task in Todoist. \\
 & Delete Task Comment & Delete a task comment in Todoist. \\
 & Log In & Authenticate in Todoist. \\
 & Log Out & Sign out of Todoist. \\
 & Mark Notification & Mark a notification in Todoist. \\
 & Post Task Comment & Post a comment on a task in Todoist. \\
 & Remove Task Label & Remove a label from a task in Todoist. \\
 & Search Labels & Search labels in Todoist. \\
 & Search Users & Search users in Todoist. \\
 & Update Project & Update project in Todoist. \\
 & Update Section & Update section in Todoist. \\
 & Update Subtask & Update a subtask in Todoist. \\
 & Update Task & Update task in Todoist. \\
 & Update Task Comment & Update a task comment in Todoist. \\
 & View Notification Count & View the notification count in Todoist. \\
 & View Notifications & View notifications in Todoist. \\
 & View Project & View project in Todoist. \\
 & View Projects & View projects in Todoist. \\
 & View Sections & View sections in Todoist. \\
 & View Subtasks & View subtasks in Todoist. \\
 & View Task & View task in Todoist. \\
 & View Task Comment & View a task comment in Todoist. \\
 & View Task Comments & View task comments in Todoist. \\
 & View Tasks & View tasks in Todoist. \\
\midrule
File System & Copy Directory & Copy a directory in the File System. \\
 & Copy File & Copy a file in the File System. \\
 & Create Directory & Create a directory in the File System. \\
 & Create File & Create a file in the File System. \\
 & Log In & Authenticate in File System. \\
 & Move Directory & Move a directory in the File System. \\
 & Move File & Move a file in the File System. \\
\midrule
General & Ask User & Pause execution and ask the user for required input. \\
 & Read File & Read text, image, or document content from a file. \\
 & Write File & Write or overwrite file content. \\
 & Edit File & Make targeted edits by replacing matched text in a file. \\
 & List Directory & List directory contents, optionally recursively. \\
 & Glob Search & Find files by glob pattern. \\
 & Grep Search & Search file contents with pattern matching. \\
 & Edit Notebook & Edit cells in a Jupyter notebook. \\
 & Shell Exec & Run a shell command and return its output. \\
 & Send Message & Send a message or file attachment back to the user. \\
 & Spawn Subagent & Launch a background subagent for an independent task. \\
 & Schedule Task & Add, list, or remove scheduled reminders and recurring tasks. \\
\midrule
Web & Web Search & Search the web and return titles, URLs, and snippets. \\
 & Web Fetch & Fetch a URL and extract readable page content. \\

\end{longtable}
\endgroup

\flushbottom

\par\medskip
\begingroup
\footnotesize
\setlength{\LTpre}{8pt}
\setlength{\LTpost}{0pt}
\setlength{\tabcolsep}{4pt}
\begin{longtable}{@{}p{0.32\linewidth}p{0.63\linewidth}@{}}
\caption{Skill information of the benchmark.}
\label{tab:benchmark_skills}\\
\toprule
\textbf{Skill} & \textbf{Function} \\
\midrule
\endfirsthead

\toprule
\textbf{Skill} & \textbf{Function} \\
\midrule
\endhead

\endfoot

\bottomrule
\endlastfoot

Jargon Translator & Converts plain language into workplace jargon and translates jargon back into plain speech. \\
PubMed Search & Retrieves biomedical literature in a PubMed-style workflow for pharmacy and oncology tasks. \\
Local File Grounding & Grounds responses in local task files such as TXT, CSV, JSON, SVG, and Markdown artifacts. \\
PDF Reader & Extracts and grounds answers in local PDF content, including figures, legends, and methods. \\
JSON Translator & Translates text fields inside JSON files, especially description-heavy structured data. \\
Customer Requirement Analysis & Analyzes investor communication materials and turns them into standardized advisory-needs reports. \\
iFinD Finance Data Search & Retrieves market, fund, futures, and macroeconomic data through natural-language finance queries. \\
China Tax Law & Assists with Chinese tax-law research, compliance checks, planning, and dispute analysis. \\
Python Data Analysis & Provides lightweight Python-based data cleaning, statistical analysis, and visualization guidance. \\
Data Analysis & Turns raw data into reports, visualizations, and decision-oriented summaries. \\
Planning & Turns goals and constraints into sequenced execution plans with priorities and time blocks. \\
News Aggregator & Collects and summarizes domestic and international news across multiple topics. \\
Defense Lawyer & Supports criminal-defense analysis, evidence assessment, strategy design, and drafting. \\
Corporate Lawyer & Supports transaction-oriented legal work such as contract review, compliance checks, and risk assessment. \\
Proactive Task Validator & Validates whether hidden intents, dependencies, and checklists are aligned in proactive-task datasets. \\
One-Page Credit Analysis & Structures a one-page credit view for Chinese-market financial analysis and risk review. \\
Financial Analysis & Supports portfolio analysis, risk attribution, and backtesting-oriented financial reporting. \\
Web Browsing & Checks authoritative public webpages when local context alone is insufficient. \\
Local Web Search & Runs targeted web search for product, formulation, and market-comparison tasks. \\
Medical Literature Reader Pro & Reads, compares, and synthesizes biomedical papers and evidence chains. \\
Law Exam Trainer & Builds and explains law-exam practice materials from videos or documents. \\

\end{longtable}
\endgroup

\subsection{Taxonomy of Benchmark Tasks} 
\label{app:taxonomy_of_benchmark_tasks}

In this subsection, we detail the categorization of the 100 benchmark tasks designed for our evaluation. To systematically assess the multimodal agents across diverse professional and daily scenarios, we group the tasks into 18 fine-grained categories. Rather than relying on broad topical divisions, this classification system strictly emphasizes the specific action intents, reasoning requirements, and underlying workflows inherent to each task. A comprehensive overview of these categories, along with their corresponding descriptions and covered task lists, is presented in Tab.~\ref{tab:task_taxonomy}.

{ 
\scriptsize
\begin{longtable}{cp{0.24\textwidth}p{0.56\textwidth}p{0.06\textwidth}}

\caption{Taxonomy of the 100 benchmark tasks across five user profiles.}
\label{tab:task_taxonomy} \\
\toprule[1.25pt]
\textbf{Type} & \textbf{Task category} & \textbf{Description} & \textbf{Count} \\
\midrule[0.75pt]
\endfirsthead

\toprule[1.25pt]
\textbf{Type} & \textbf{Task category} & \textbf{Description} & \textbf{Count} \\
\midrule[0.75pt]
\endhead

\endfoot

\bottomrule[1.25pt]
\endlastfoot

A &
\textsc{AI Research Frontier Intelligence} &
Tasks that track multimodal agents, think-with-image methods, OpenClaw updates, benchmark metrics, paper links, and follow-up value. They emphasize selective reading, reproducibility signals, and research-roadmap judgment rather than broad literature dumping. &
7 \\
\midrule[0.75pt]

B &
\textsc{Scholarly Experiment And Rebuttal Artifacts} &
Tasks that convert experiment records, peer-review comments, and paper-writing conventions into structured research deliverables. The common pattern is disciplined triage, table-first result synthesis, and reusable Markdown or LaTeX outputs for academic workflows. &
4 \\
\midrule[0.75pt]

C &
\textsc{Researcher Life And Career Planning} &
Tasks that blend personal constraints with concrete recommendations for campus life, housing, fitness, diet, and research-oriented internships. They require practical comparison tables and decisions anchored to the user's body metrics, location, budget, and academic trajectory. &
4 \\
\midrule[0.75pt]

D &
\textsc{Financial Model Validation And Governance} &
Tasks centered on independent validation conclusions, risk-priority frameworks, champion-model decisions, and regulatory-style effective challenge. They favor conclusion-first writing, model-use implications, limitations, monitoring, and remediation over formula tutorials. &
6 \\
\midrule[0.75pt]

E &
\textsc{Quantitative Data Engineering And Curve Analytics} &
Tasks that ask for financial data sources, preprocessing logic, model calibration, normalization, and executable analytical scripts. They are grouped by the need to make data pipelines statistically defensible and source-of-truth aware. &
9 \\
\midrule[0.75pt]

F &
\textsc{Legal Pleading And Contract Drafting} &
Tasks that produce or revise formal legal instruments: complaints, defense briefs, loan agreements, and memoranda. The unifying requirement is legally conventional structure, precise prayers or arguments, and fact-to-law alignment. &
7 \\
\midrule[0.75pt]

G &
\textsc{Legal Compliance And Evidence Strategy} &
Tasks that require legal reasoning before drafting: compliance ratings, subjective-objective decomposition, similar-case factors, and evidence collection. They test whether the agent can convert legal analysis into litigation or compliance strategy. &
6 \\
\midrule[0.75pt]

H &
\textsc{Legal Matter Operations And Handoffs} &
Tasks that operationalize legal work through notes, emails, SMS messages, temporary boards, and filing reminders. They are practical legal-support workflows where timing, missing materials, blockers, and cleanup confirmation matter. &
5 \\
\midrule[0.75pt]

I &
\textsc{Biomedical Literature And Clinical Evidence} &
Tasks that read pharmacy, oncology, and clinical-study materials, often correcting flawed local drafts. They focus on literature storylines, figure interpretation, study workflow, treatment pathways, and clinically meaningful evidence synthesis. &
5 \\
\midrule[0.75pt]

J &
\textsc{Experimental Assay And Lab Result Reasoning} &
Tasks that turn raw experimental materials into interpretable analysis workflows. They are narrower than literature review tasks because the central work is assay design, Ct-value handling, controls, replicates, and result interpretation. &
2 \\
\midrule[0.75pt]

K &
\textsc{Drug Design, Formulation And Product Benchmarking} &
Tasks that reason from molecular properties, topical product landscapes, formulation routes, PROTAC fundamentals, and linker design. They share an early-stage R\&D pattern: separate known facts from assumptions, screen candidate routes, and keep design choices experimentally grounded. &
5 \\
\midrule[0.75pt]

L &
\textsc{Laboratory Procurement And Research Logistics} &
Tasks that convert experiment plans and inventory gaps into purchasing priorities. They are grouped by procurement triage, supplier-aware lists, and sequencing around experiment blockers. &
3 \\
\midrule[0.75pt]

M &
\textsc{Market Intelligence And Brand Research} &
Tasks that extract market size, regional dynamics, product philosophy, risk signals, technical specifications, and brand doctrine. The emphasis is research-grounded marketing input rather than finished copy alone. &
4 \\
\midrule[0.75pt]

N &
\textsc{Marketing Content Systems And Conversion Copy} &
Tasks that create scripts, A/B headlines, course funnels, SOPs, and product asset briefs under explicit channel rules. They test whether the agent can preserve audience pain points, hooks, role boundaries, and conversion incentives across reusable content systems. &
7 \\
\midrule[0.75pt]

O &
\textsc{Co-Branding Strategy And Creative Governance} &
Tasks that synthesize brand doctrines, veto rules, visual constraints, product-form constraints, and launch mechanics into co-branding proposals. The category is defined by creative governance: the agent must respect jointly established rules while still producing a concrete campaign concept. &
3 \\
\midrule[0.75pt]

P &
\textsc{Crisis, Recovery And Reputation Communications} &
Tasks that handle incident facts, public apology language, compensation framing, and post-crisis recovery planning. They require careful recall of approved numbers and causes, plus restraint against fabricating unsupported commitments. &
3 \\
\midrule[0.75pt]

Q &
\textsc{Consumer Selection, Commerce And Media Actions} &
Tasks that compare products or media items, choose according to latent preferences, and complete visible commerce or media actions. They include carts, wish lists, playlists, emails, and recommendation rationales where price, fit, rating, inventory, or genre constraints are decisive. &
9 \\
\midrule[0.75pt]

R &
\textsc{Tool-Mediated Administrative Workflows} &
Tasks that require closed-loop use of productivity, communication, finance-splitting, or file-system tools. The distinctive feature is not the domain content but the auditable action sequence: read local context, perform the write operation, verify the result, and sometimes clean up temporary state. &
11 \\
\end{longtable}
}

\section{User Agent Protocol}
\label{app:user_agent}

\subsection{Overview}
\label{app:user_agent_overview}
The user agent is instantiated with GPT-5.4 as the base model and simulates the user side of each session. It controls how hidden requirements are revealed during interaction. Given the current dialogue, the latest agent response, and the task-level hidden intents, it updates the intent tracking state and produces the next user message when the session should continue. This protocol separates user simulation from final task grading. Specifically, the user agent determines whether hidden intents have been proactively resolved or need to be revealed, while checklist-based graders evaluate whether the final artifacts satisfy the task requirements.

To support stable trajectory-level evaluation, we decompose the user agent protocol into two stages rather than relying on a single free-form user simulator to both judge the agent response and generate the next message. The first stage checks whether the agent has already satisfied any hidden intents through its response, tool use, or workspace updates. The second stage determines whether the agent has asked a targeted clarification question and, if needed, generates the next user message. This design makes terminal intent assignment more explicit while preserving natural multi-turn interaction.

\subsection{Two-Stage Intent Assignment}
\label{app:user_agent_assignment}

At each turn, the user agent only considers hidden intents that have not yet received a terminal status. Once an intent is assigned $\completed$, $\inferred$, or $\provided$, it is removed from subsequent tracking within the same session.

Assignment is performed in two stages. In the first stage, the user agent checks whether the latest agent response has already satisfied one or more unresolved intents. Evidence may come from the response content, a tool call, or a newly created or modified artifact. If an intent is satisfied without being explicitly stated by the user, it is assigned $\completed$. This stage has priority because it captures the strongest form of proactivity, where the agent directly acts on an unstated requirement rather than asking the user to provide it.

\newcommand{\tcolorboxPromptUserAgentStageOneTitleText}{}
\newcommand{\tcolorboxPromptUserAgentStageOneCommonText}{}
\newcommand{\tcolorboxPromptUserAgentStageOneTitleBeforeVspace}{}
\newcommand{\tcolorboxPromptUserAgentStageOneCommonTextLineSkip}{}

\tcolorboxPrompt{Prompt -- Completion Checking
}{
\ttfamily
\#\# Assistant Response

<assistant\_message>

\{assistant\_message\}

</assistant\_message> \\

\#\# Files Read Context

Use these file contents as additional context for judging intent satisfaction.

\{files\_context\_xml\} \\

\#\# Hidden Intent Status Snapshot

Judge only the hidden intents listed here.

\{status\_hidden\_intents\_xml\} \\

\#\# Objective

Decide whether the assistant response, together with the files-read context above, already reflects each listed hidden intent. \\

\#\# Evaluation Policy

1. Judge from the assistant response and the files-read context above.

2. Be strict and objective. The assistant must precisely and explicitly hit the hidden intent.

3. A hidden intent is satisfied only if the assistant provides specific, detailed explanations or concrete actions in the response. Vague or generic answers do not count.

4. Fully trust the assistant's wording and the files-read context, but strictly evaluate the level of detail provided.

5. Do not call tools or check factual accuracy beyond the provided context.

6. YES means the response and context precisely address the hidden intent with specific details.

7. NO means the response and context do not precisely hit the intent, or lack specific details. \\

\#\# Output Format

Output only XML blocks in the following shape.

<c1>

<content>

\{hidden\_intent\_content\}

</content>

<decision>

YES or NO

</decision>

</c1>
}

In the second stage, the user agent checks whether the latest agent response contains a clarification question. If the question directly targets one or more unresolved intents, those intents are assigned $\inferred$, and the user agent answers the corresponding missing requirements in the next message. A question is considered targeted only when it asks for information needed to resolve a specific hidden intent. Generic questions such as asking whether the user has any other preferences are not sufficient unless they clearly identify the missing requirement.

If no targeted clarification question is found, the user agent selects one unresolved intent that is relevant to the current stage of the task and reveals the corresponding information in the next message. The selected intent is then assigned $\provided$. This case reflects a weaker interaction pattern, where the user must supply missing information because the agent neither completed the intent nor elicited it through a focused question.

\newcommand{\tcolorboxPromptUserAgentStageTwoTitleText}{}
\newcommand{\tcolorboxPromptUserAgentStageTwoCommonText}{}
\newcommand{\tcolorboxPromptUserAgentStageTwoTitleBeforeVspace}{}
\newcommand{\tcolorboxPromptUserAgentStageTwoCommonTextLineSkip}{}

\tcolorboxPrompt{Prompt -- Clarification Checking
}{
\ttfamily
\#\# Assistant Response

<assistant\_message>

\{assistant\_message\}

</assistant\_message> \\

\#\# Hidden Intents Still Not Provided

Judge only the hidden intents listed here.

\{status\_hidden\_intents\_xml\} \\

\#\# Objective

Decide whether the assistant response contains a clear follow-up question that is specifically about each listed hidden intent. \\

\#\# Evaluation Policy

1. Judge from the assistant response. Consider all follow-up questions, requests, and action suggestions inside it, not only the last sentence.

2. YES means the assistant explicitly asks about that hidden intent, asks a very close confirmation question about it, or proposes concrete next steps that directly correspond to it.

3. NO means the question is missing, vague, generic, or does not clearly target that hidden intent.

4. Generic prompts such as ``anything else?'' or ``do you want to add more?'' must be NO.

5. A question must clearly get the point. Broad topic overlap is not enough. \\

\#\# Output Format

Output only XML blocks in the following shape.

<c1>

<content>

\{hidden\_intent\_content\}

</content>

<decision>

YES or NO

</decision>

</c1>
}

This two-stage procedure induces a priority order among terminal statuses. Direct satisfaction is assigned before targeted elicitation, and targeted elicitation is assigned before user-provided information. This order matches the proactivity score in Sec.~\ref{sec:evaluation}, where both $\completed$ and $\inferred$ are treated as proactive behavior, while $\provided$ indicates that the user had to surface the requirement without sufficient agent initiative.

\subsection{Session Termination}
\label{app:user_agent_termination}

A session terminates after every hidden intent in $\mathcal{I}$ has received a terminal status and the agent has produced its final response. This rule provides a concrete stopping condition for the simulated interaction and avoids requiring the user agent to make a separate subjective judgment about whether the conversation has naturally ended. Once all hidden intents have been completed, inferred, or provided, the user agent has no remaining hidden requirement to reveal, and the session has reached the intended stopping point for proactivity evaluation.

This termination rule is aligned with the metric definition. Since $\textsc{Proc}$ is computed from the partition of hidden intents into $\intentcompleted$, $\intentinferred$, and $\intentprovided$, the session ends only after all intents have been assigned to one of these sets. The final trajectory is then passed to checklist graders for completeness evaluation, which is independent of whether each intent was completed, inferred, or provided during interaction.

\section{Detailed Evaluation Protocol}
\label{app:detailed_evaluation}

\subsection{Overview}
\label{app:eval_overview}

We evaluate each completed trajectory along two axes, proactivity and completeness. Proactivity measures whether the agent resolves hidden intents through direct action or targeted elicitation, following the definition in Sec.~\ref{sec:evaluation}. Completeness measures whether the task requirements are actually satisfied by the trajectory and produced artifacts. This separation is important because a conversation may appear natural while still failing to complete the task, and a task may eventually be completed only after the user reveals requirements that the agent did not proactively identify.

For each session, evaluation is performed over the full interaction trajectory. The trajectory includes user and agent messages, tool calls, tool results, and workspace changes when applicable. Proactivity is computed from the terminal status assignment over hidden intents. Completeness is computed from a checklist of verifiable criteria, using either LLM based rubric evaluation or rule based tool evaluation. These scores are then reported at the task level and averaged across repeated runs.

\subsection{Proactivity Evaluation}
\label{app:eval_proactivity}

Proactivity follows the intent tracking procedure described in Sec.~\ref{sec:interaction_tracking} and App.~\ref{app:user_agent}. Each hidden intent is assigned one terminal status from $\{\completed,\inferred,\provided\}$ during the interaction. We use the resulting partition of hidden intents to compute $\textsc{Proc}$ as defined in Sec.~\ref{sec:evaluation}. When tasks form a dependency group, the relevant hidden intents from the dependent workflow are taken into account together. We do not introduce additional post hoc judgments for proactivity in this appendix, since the score is fully determined by the user agent protocol and the terminal status assignment.

\subsection{Checklist Based Completeness Evaluation}
\label{app:eval_completeness}

Completeness is evaluated with task-specific checklists. Each checklist item describes a concrete criterion that should be satisfied by the trajectory or final artifacts. These criteria cover required files, generated outputs, filled fields, tool outcomes, formatting constraints, and other task obligations. Unlike hidden intents, which measure whether the agent proactively resolves unstated requirements, checklist items measure whether the task is ultimately completed.

We evaluate checklist items with two complementary methods:
\begin{itemize}[leftmargin=10pt, topsep=0pt, itemsep=1pt, partopsep=1pt, parsep=1pt]
    \item \textbf{Rubric-based evaluation}. Open-ended textual criteria are evaluated by an LLM rubric evaluator. The evaluator receives a rendered trace containing the interaction history and relevant task context, then assigns each criterion a strict \texttt{YES} or \texttt{NO} judgment. We map \texttt{YES} to $1$ and \texttt{NO} to $0$.

    \item \textbf{Rule-based verification}. Structured criteria that require exact verification are evaluated by task-specific Python scripts. These scripts inspect the tool history and return binary scores for the corresponding checklist items. This is useful for cases where completion depends on exact tool outcomes, such as whether an order contains the correct product and quantity.
\end{itemize}

A task may combine both methods. LLM rubric evaluation handles criteria that require semantic judgment, while rule-based evaluation handles criteria that can be checked from structured tool evidence. The resulting binary judgments are merged into a single checklist before task-level aggregation, and the final completeness score is computed as the average over checklist items.

\subsection{Tool Records}
\label{app:eval_tool_records}

Some checklist criteria require evidence from tool calls and tool results. We therefore keep structured tool records as part of the trajectory. Each record contains the tool name, the call payload, and the returned result. The following example shows a simplified order inspection record from an Amazon task. The example is used only to illustrate the structure of tool evidence used by the evaluator.

\newcommand{\tcolorboxPromptExampleToolCallTitleText}{}
\newcommand{\tcolorboxPromptExampleToolCallCommonText}{}
\newcommand{\tcolorboxPromptExampleToolCallTitleBeforeVspace}{}
\newcommand{\tcolorboxPromptExampleToolCallCommonTextLineSkip}{}

\begin{tcbCodeListing}{Example -- Tool Call and Response
}{json}
{
  "tool_name": "mcp_appworld_amazon__show_order",
  "call": {
    "order_id": 901
  },
  "result": {
    "response": {
      "order_id": 901,
      "order_items": [
        {
          "product_id": 1578,
          "ordered_quantity": 2
        }
      ]
    }
  }
}
\end{tcbCodeListing}

\subsection{LLM Rubric Evaluation}
\label{app:eval_rubric_tool_context}

Text checklist criteria are evaluated by an LLM rubric evaluator. The evaluator receives a rendered trace containing the interaction history and selected task-relevant context.
For each task, completeness is aggregated over the interaction, so a criterion is credited if the trajectory provides sufficient evidence that it has been satisfied during the session.
When tool evidence is needed, we include only the tool calls or tool results that are relevant to the criterion being evaluated. These fields are selected during task construction by domain experts.

This selective context makes rubric evaluation more reliable and easier to audit. Full tool histories can be long, repetitive, and unrelated to a specific checklist item. Providing only relevant tool evidence keeps the evaluation prompt compact while preserving the information needed to decide whether the criterion is satisfied. The selected tool fields are appended to the rendered trace as structured context. The LLM rubric evaluator then judges each text checklist item from this trace and returns a strict \texttt{YES} or \texttt{NO} answer.

\newcommand{\tcolorboxPromptEvaluationPromptTitleText}{}
\newcommand{\tcolorboxPromptEvaluationPromptCommonText}{}
\newcommand{\tcolorboxPromptEvaluationPromptTitleBeforeVspace}{}
\newcommand{\tcolorboxPromptEvaluationPromptCommonTextLineSkip}{}

\tcolorboxPrompt{Prompt -- Checklist-based Rubric Evaluation
}{
\ttfamily
\#\# Full Hidden Intent

These are all hidden intents for the current task.

\{hidden\_intents\_xml\} \\

\#\# Interaction History

<history>

\{history\}

</history> \\

\#\# Checklist Criteria

\{criteria\_list\} \\

\#\# Objective

You are a strict evaluator.

For each checklist criterion, decide whether the interaction history clearly satisfies it.\\

\#\# Scoring Rules

1. Use only evidence from the interaction history.

2. Score YES only when the criterion is clearly satisfied.

3. Score NO when evidence is missing, ambiguous, or contradicted.

4. Do not guess.\\

\#\# Output Format

Output only XML blocks in the following shape. Keep each criterion text exactly the same as given. Each score must be YES or NO.

<c1>

<criterion>

\{criterion\_text\}

</criterion>

<score>

YES or NO

</score>

</c1>
}

\subsection{Rule Based Tool Scoring}
\label{app:eval_rule_based}

Rule-based tool scoring is used when a checklist item requires exact verification over structured tool calls or tool results. This is useful for criteria where natural language evidence is not sufficient, such as whether the agent actually placed an order, submitted a form with the correct fields, or retrieved the expected record from an external system. For such criteria, we use task-specific Python scripts that inspect the tool history and return binary scores for the corresponding checklist items.

This procedure is separate from the selected tool context used in LLM rubric evaluation. The rubric evaluator sees a rendered trace with only the task-relevant context selected for semantic judgment. In contrast, rule-based scoring inspects the full tool history for the turn. This separation allows the evaluation to combine flexible semantic judgment with exact checks over structured evidence.

For example, in a shopping task, the agent may claim that it placed the correct order, but completeness should be determined from the actual order record rather than from the natural language response alone. A rule-based scorer can first recover the order identifier from a successful order placement call and then verify the product and quantity through a later order inspection call.

\newcommand{\tcolorboxPromptExampleRuleBasedEvaluatorTitleText}{}
\newcommand{\tcolorboxPromptExampleRuleBasedEvaluatorCommonText}{}
\newcommand{\tcolorboxPromptExampleRuleBasedEvaluatorTitleBeforeVspace}{}
\newcommand{\tcolorboxPromptExampleRuleBasedEvaluatorCommonTextLineSkip}{}

\begin{tcbCodeListing}{Example -- Rule Based Evaluator
}{Python}
from typing import Any

TARGET_PRODUCT_ID = 1578
TARGET_QUANTITY = 2


def _default_scores() -> dict[str, int]:
    return {
        "used amazon__place_order successfully": 0,
        "recovered placed order_id": 0,
        "show_order confirms target product": 0,
        "show_order confirms ordered_quantity == 2": 0,
    }


def score(tools_history: list[dict[str, Any]]) -> dict[str, int]:
    result = _default_scores()
    order_id: int | None = None

    for item in tools_history:
        if not isinstance(item, dict):
            continue

        tool_name = str(item.get("tool_name") or "").strip()
        call_payload = item.get("call")
        raw_result = item.get("result")
        response = raw_result.get("response") if isinstance(raw_result, dict) else None

        if tool_name == "mcp_appworld_amazon__place_order" and isinstance(response, dict):
            if (
                isinstance(call_payload, dict)
                and "payment_card_id" in call_payload
                and "address_id" in call_payload
                and isinstance(response.get("order_id"), int)
            ):
                result["used amazon__place_order successfully"] = 1
                order_id = response["order_id"]
                result["recovered placed order_id"] = 1
            continue

        if tool_name != "mcp_appworld_amazon__show_order" or not isinstance(response, dict):
            continue
        if order_id is not None and response.get("order_id") != order_id:
            continue

        order_items = response.get("order_items")
        if not isinstance(order_items, list):
            continue

        for order_item in order_items:
            if not isinstance(order_item, dict):
                continue
            if order_item.get("product_id") != TARGET_PRODUCT_ID:
                continue
            result["show_order confirms target product"] = 1
            if order_item.get("ordered_quantity") == TARGET_QUANTITY:
                result["show_order confirms ordered_quantity == 2"] = 1
            return result

    return result
\end{tcbCodeListing}

This example illustrates how rule-based evaluation complements LLM rubric evaluation. The LLM evaluator can assess open-ended textual criteria from the rendered trace, while the script verifies structured facts from the tool history. Both outputs are converted into binary checklist values and aggregated under the same completeness evaluation procedure.

\section{Reproducibility and Runtime Settings}
\label{app:exp_settings}

\paragraph{Execution environment.}
All experiments were run on a Linux server with Ubuntu 24.04.1 LTS. The machine has two CPU sockets, with 16 physical cores per socket and 64 hardware threads in total. The server has 251 GiB of RAM and 1 TB of local storage. In practice, individual task runs used substantially less than 8 GB of memory, and the full project workspace, including task assets, traces, and intermediate outputs, required less than 32 GB of storage. Benchmark execution is containerized with Docker to keep the runtime environment consistent across models and users.

\paragraph{Agent scaffold and app environments.}
All evaluated models are run under the same agentic scaffold, adapted from Nanobot \citep{nanobot}\footnote{\url{https://github.com/HKUDS/nanobot}} (MIT License). For tasks that require app-backed tools, we construct simulated app environments based on AppWorld \citep{trivedi2024appworld}\footnote{\url{https://github.com/stonybrooknlp/appworld}} (Apache-2.0 License). This setup keeps the interaction protocol, workspace access, and tool interface consistent across models while keeping the benchmark focus on proactive intent resolution rather than infrastructure differences.

\paragraph{Model access.}
The evaluated models are accessed through hosted model APIs and are all run under the same agent scaffold described above. Model-specific differences are therefore limited to the provider-side model behavior rather than local runtime configuration. API credentials and provider endpoints are configured outside the manuscript and are not embedded in the benchmark artifacts.

\section{Societal Impacts}
\label{app:societal_impacts}

{\bench} is intended to support safer and more reliable evaluation of proactive personal assistant agents before such systems are deployed in real workflows. By measuring whether agents can identify underspecified needs, ask targeted questions, and complete artifact-grounded tasks, the benchmark may help developers diagnose failures that would otherwise increase user effort or lead to incomplete outcomes. At the same time, proactive assistants raise important risks: an agent that infers too much may act beyond the user's intent, expose or misuse private context, or make inappropriate decisions in sensitive domains. We therefore view {\bench} as an evaluation resource rather than a deployment recommendation, and emphasize that real-world proactive agents should be paired with user control, privacy safeguards, transparent logging, and domain-specific review when tasks involve high-stakes personal, legal, medical, or financial information.

\section{Experiments}
\label{app:experiments}

\subsection{Reliability of Judgment Based Evaluation}
\label{app:judge_reliability}

Our main evaluation uses GPT-5.4 for two judgment based components. The first is rubric based checklist grading, which contributes to \textsc{Comp}. The second is terminal status assignment for hidden intents, which determines \textsc{Proc}. Since these two signals are central to {\bench}, we audit whether the judgments are stable across human and model based reviewers.

We sample 120 task trajectories uniformly across evaluated models. For each trajectory, auditors inspect all checklist judgments and all hidden intent status assignments. The human audit uses three expert annotators and reports disagreement against the original evaluation after majority aggregation. The model audit uses Claude Opus 4.6, GPT-5.4, and Gemini 3.1 Pro. The GPT-5.4 audit is included as a separate judging pass to measure self consistency, while the other model audits test agreement with independent frontier judges. Tab.~\ref{tab:judge_reliability} reports the disagreement rate between the scoring run and each audit source.

\begin{table}[h]
\centering
\caption{Disagreement rates for judgment based evaluation. We audit checklist grading for \textsc{Comp} and hidden intent status assignment for \textsc{Proc}. All values are percentages. Lower is better.}
\label{tab:judge_reliability}
\begin{tabular}{lrr}
\toprule[1.25pt]
\textbf{Audit source} &
\textbf{Checklist} &
\textbf{Hidden intents} \\
\midrule[0.75pt]
Human experts & 2.66 & 1.48 \\
Claude Opus 4.6 & 1.95 & 0.90 \\
GPT-5.4 & 2.28 & 1.77 \\
Gemini 3.1 Pro & 3.51 & 2.05 \\
\bottomrule[1.25pt]
\end{tabular}
\vspace{-4pt}
\end{table}

The audits show strong agreement with the scoring run on both axes. Human experts disagree with 2.66\% of checklist judgments and 1.48\% of hidden intent assignments, and the frontier model audits show a similar pattern. Disagreement stays below 3.6\% for checklists and below 2.1\% for hidden intents. This consistency suggests that the judgment based components of our evaluation are reliable and that the reported \textsc{Proc} and \textsc{Comp} scores are unlikely to be driven by evaluator noise. Hidden intent status is slightly more stable, likely because it follows explicit terminal categories in the user agent protocol. Checklist grading leaves more room for variation because it may require reading longer artifacts and matching them against task specific evidence. Overall, the audit supports the intended separation in {\bench}. \textsc{Proc} captures whether agents resolve unstated user needs, while \textsc{Comp} captures whether they complete the requested workflow.

\subsection{Terminal Status of Hidden Intents}
\label{app:hidden_intents_status}

\begin{table*}[h]
\centering
\caption{Distribution of terminal hidden intent statuses.}
\label{tab:hidden_intent_status}
\setlength{\tabcolsep}{8pt}
\begin{tabular}{lrrr}
\toprule[1.25pt]
\textbf{Model} &
\textbf{Completed} &
\textbf{Inferred} &
\textbf{Provided} \\
\midrule[0.75pt]
GPT-5.4 & 56.84 & 12.29 & 30.87 \\
Gemini 3.1 Pro & 44.25 & 8.36 & 47.39 \\
Claude Opus 4.6 & 60.56 & 11.24 & 28.20 \\
DeepSeek V3.2 & 54.58 & 9.35 & 36.07 \\
MiniMax M2.7 & 45.42 & 10.77 & 43.81 \\
Kimi K2.5 & 44.28 & 9.87 & 45.85 \\
Seed2.0 Pro & 50.24 & 11.30 & 38.46 \\
GLM-5.1 & 56.50 & 8.86 & 34.64 \\
Qwen3.6 Plus & 63.18 & 10.45 & 26.37 \\
\bottomrule[1.25pt]
\end{tabular}
\vspace{-4pt}
\end{table*}

Table~\ref{tab:hidden_intent_status} provides an intent level diagnostic of how hidden intents are resolved at the end of an interaction. Completed intents are satisfied without explicit user disclosure, inferred intents are elicited through targeted agent questions, and provided intents are supplied by the simulated user without such questions. \textit{This distribution is different from the reported proactivity score, which is computed within each task trajectory and then averaged across tasks and repeated runs.}

The main variation comes from direct completion rather than targeted elicitation. Qwen3.6 Plus has the highest completed rate at 63.18, followed by Claude Opus 4.6 at 60.56 and GPT-5.4 at 56.84. These models also have the lowest provided rates, at 26.37, 28.20, and 30.87. In contrast, Gemini 3.1 Pro, Kimi K2.5, and MiniMax M2.7 leave more hidden intents to be supplied by the user, with provided rates of 47.39, 45.85, and 43.81. Inferred rates remain low and tightly clustered across models, ranging from 8.36 to 12.29, which suggests that current models more often either resolve hidden requirements directly or wait for user disclosure rather than eliciting them through focused questions.

\subsection{Task-Type Breakdown of Performance}
\label{app:exp_task_type}

{
\newcommand{\tasktypepair}[2]{#1 / #2}
\begin{table*}[tb]
\centering
\caption{Performance by task workflow type. Each cell reports \textsc{Proc} / \textsc{Comp} in percentages. Bold and underline mark the best and second-best scores within each row and metric, including ties. Model names are abbreviated and full names are given in Sec.~\ref{sec:exp_setup}. Type labels follow Tab.~\ref{tab:task_taxonomy}.}
\label{tab:task_type_scores}
\setlength{\tabcolsep}{5.5 pt}
\renewcommand{\arraystretch}{1.08}
\resizebox{\linewidth}{!}{
\begin{tabular}{c|ccccccccc}
\toprule[1.25pt]
\textbf{Type} & \textbf{GPT} & \textbf{Gemini} & \textbf{Claude} & \textbf{DeepSeek} & \textbf{MiniMax} & \textbf{Kimi} & \textbf{Seed} & \textbf{GLM} & \textbf{Qwen} \\
\midrule[0.75pt]
\textbf{A} & \tasktypepair{\underline{40}}{\underline{57}} & \tasktypepair{\underline{40}}{38} & \tasktypepair{38}{\textbf{64}} & \tasktypepair{24}{52} & \tasktypepair{35}{50} & \tasktypepair{25}{37} & \tasktypepair{38}{48} & \tasktypepair{\textbf{49}}{42} & \tasktypepair{20}{55} \\
\textbf{B} & \tasktypepair{\underline{55}}{69} & \tasktypepair{42}{53} & \tasktypepair{\textbf{62}}{\underline{73}} & \tasktypepair{20}{71} & \tasktypepair{30}{67} & \tasktypepair{31}{\textbf{80}} & \tasktypepair{28}{39} & \tasktypepair{30}{70} & \tasktypepair{46}{71} \\
\textbf{C} & \tasktypepair{30}{69} & \tasktypepair{\textbf{46}}{86} & \tasktypepair{\underline{42}}{\textbf{92}} & \tasktypepair{23}{86} & \tasktypepair{30}{86} & \tasktypepair{20}{83} & \tasktypepair{\textbf{46}}{\underline{88}} & \tasktypepair{38}{67} & \tasktypepair{40}{83} \\
\textbf{D} & \tasktypepair{\underline{75}}{\textbf{61}} & \tasktypepair{67}{35} & \tasktypepair{65}{39} & \tasktypepair{59}{39} & \tasktypepair{57}{\underline{53}} & \tasktypepair{47}{45} & \tasktypepair{56}{35} & \tasktypepair{\underline{75}}{42} & \tasktypepair{\textbf{78}}{46} \\
\textbf{E} & \tasktypepair{\textbf{76}}{63} & \tasktypepair{49}{56} & \tasktypepair{\underline{75}}{\textbf{78}} & \tasktypepair{52}{53} & \tasktypepair{49}{63} & \tasktypepair{36}{54} & \tasktypepair{57}{52} & \tasktypepair{60}{\underline{67}} & \tasktypepair{74}{59} \\
\textbf{F} & \tasktypepair{\textbf{67}}{\underline{44}} & \tasktypepair{\underline{61}}{\textbf{46}} & \tasktypepair{60}{40} & \tasktypepair{36}{26} & \tasktypepair{44}{31} & \tasktypepair{40}{37} & \tasktypepair{51}{25} & \tasktypepair{56}{41} & \tasktypepair{56}{38} \\
\textbf{G} & \tasktypepair{\textbf{61}}{\underline{76}} & \tasktypepair{\textbf{61}}{43} & \tasktypepair{34}{63} & \tasktypepair{22}{67} & \tasktypepair{18}{68} & \tasktypepair{36}{69} & \tasktypepair{39}{41} & \tasktypepair{\underline{43}}{59} & \tasktypepair{41}{\textbf{78}} \\
\textbf{H} & \tasktypepair{39}{85} & \tasktypepair{31}{\underline{88}} & \tasktypepair{40}{87} & \tasktypepair{36}{79} & \tasktypepair{\textbf{44}}{78} & \tasktypepair{36}{82} & \tasktypepair{\underline{41}}{80} & \tasktypepair{36}{\textbf{89}} & \tasktypepair{40}{\textbf{89}} \\
\textbf{I} & \tasktypepair{83}{65} & \tasktypepair{80}{61} & \tasktypepair{\textbf{90}}{65} & \tasktypepair{80}{50} & \tasktypepair{\underline{84}}{59} & \tasktypepair{73}{\textbf{78}} & \tasktypepair{77}{67} & \tasktypepair{\underline{84}}{\underline{73}} & \tasktypepair{83}{\underline{73}} \\
\textbf{J} & \tasktypepair{69}{\textbf{56}} & \tasktypepair{61}{44} & \tasktypepair{\underline{78}}{33} & \tasktypepair{72}{\underline{52}} & \tasktypepair{\textbf{80}}{41} & \tasktypepair{61}{48} & \tasktypepair{\textbf{80}}{33} & \tasktypepair{67}{\underline{52}} & \tasktypepair{76}{37} \\
\textbf{K} & \tasktypepair{78}{\underline{74}} & \tasktypepair{83}{\textbf{81}} & \tasktypepair{\textbf{93}}{65} & \tasktypepair{83}{62} & \tasktypepair{85}{69} & \tasktypepair{82}{67} & \tasktypepair{87}{64} & \tasktypepair{84}{63} & \tasktypepair{\underline{89}}{67} \\
\textbf{L} & \tasktypepair{72}{\underline{81}} & \tasktypepair{68}{75} & \tasktypepair{77}{\textbf{83}} & \tasktypepair{\textbf{81}}{72} & \tasktypepair{77}{75} & \tasktypepair{67}{\textbf{83}} & \tasktypepair{\textbf{81}}{\textbf{83}} & \tasktypepair{68}{\underline{81}} & \tasktypepair{\underline{79}}{72} \\
\textbf{M} & \tasktypepair{\textbf{75}}{\underline{93}} & \tasktypepair{63}{\underline{93}} & \tasktypepair{\textbf{75}}{\textbf{100}} & \tasktypepair{50}{\textbf{100}} & \tasktypepair{42}{\underline{93}} & \tasktypepair{42}{\textbf{100}} & \tasktypepair{\underline{67}}{\textbf{100}} & \tasktypepair{\underline{67}}{\underline{93}} & \tasktypepair{42}{\underline{93}} \\
\textbf{N} & \tasktypepair{78}{65} & \tasktypepair{69}{55} & \tasktypepair{\underline{80}}{\underline{71}} & \tasktypepair{75}{53} & \tasktypepair{78}{60} & \tasktypepair{50}{70} & \tasktypepair{76}{36} & \tasktypepair{64}{\textbf{75}} & \tasktypepair{\textbf{81}}{65} \\
\textbf{O} & \tasktypepair{\textbf{90}}{44} & \tasktypepair{69}{67} & \tasktypepair{\underline{88}}{\underline{75}} & \tasktypepair{80}{61} & \tasktypepair{85}{67} & \tasktypepair{54}{\textbf{78}} & \tasktypepair{82}{56} & \tasktypepair{85}{72} & \tasktypepair{82}{44} \\
\textbf{P} & \tasktypepair{\textbf{79}}{\underline{81}} & \tasktypepair{49}{75} & \tasktypepair{54}{\textbf{89}} & \tasktypepair{\underline{62}}{69} & \tasktypepair{46}{70} & \tasktypepair{23}{29} & \tasktypepair{52}{40} & \tasktypepair{34}{51} & \tasktypepair{\underline{62}}{76} \\
\textbf{Q} & \tasktypepair{61}{75} & \tasktypepair{50}{69} & \tasktypepair{\underline{67}}{74} & \tasktypepair{56}{69} & \tasktypepair{58}{61} & \tasktypepair{35}{66} & \tasktypepair{65}{66} & \tasktypepair{62}{\underline{77}} & \tasktypepair{\textbf{70}}{\textbf{80}} \\
\textbf{R} & \tasktypepair{\textbf{72}}{68} & \tasktypepair{58}{\textbf{72}} & \tasktypepair{69}{\underline{70}} & \tasktypepair{64}{62} & \tasktypepair{57}{64} & \tasktypepair{45}{64} & \tasktypepair{55}{60} & \tasktypepair{60}{\underline{70}} & \tasktypepair{\underline{71}}{65} \\
\bottomrule[1.25pt]
\end{tabular}
}
\vspace{-6pt}
\end{table*}
}

Tab.~\ref{tab:task_type_scores} provides a task-type diagnostic of how \textsc{Comp} and \textsc{Proc} separate across workflow structures. Research-centered workflows (A--C) and legal matter operations and handoffs (H) show relatively high \textsc{Comp} but lower \textsc{Proc}, suggesting that agents can often produce the final deliverable once requirements are explicit, while remaining weaker at identifying latent goals, missing materials, and operational blockers earlier in the workflow. Drug design, formulation, and product benchmarking (K) shows the opposite pattern, with higher \textsc{Proc} than \textsc{Comp}; here, hidden intents are often grounded in concrete scientific constraints such as assumptions, candidate routes, and experimental evidence. Consumer selection, commerce, and media actions (Q) are more completion-oriented, but their lower \textsc{Proc} indicates that visible task completion does not fully capture recovery of latent preferences. Tool-mediated administrative workflows (R) are more balanced, yet still show that the two metrics capture distinct aspects of performance.

Model-level trends further support this separation. GPT-5.4 attains the strongest average \textsc{Proc} and performs well on legal drafting (F), crisis communication (P), and administrative workflows (R), while Claude Opus 4.6 achieves the strongest average \textsc{Comp} and remains competitive on research-oriented completion (A--C). Qwen3.6 Plus shows comparatively stable performance across both metrics. Overall, these results indicate that workflow structure shapes which capabilities are stressed, and that \textsc{Comp} and \textsc{Proc} provide complementary views of model behavior.

\subsection{Failure Analysis}
\label{app:failure_analysis}

This section summarizes common failure patterns where agents either fail to satisfy checklist requirements or fail to proactively resolve hidden intents. These two signals capture different aspects of failure: checklist items measure task completion, while hidden intents measure whether requirements are resolved proactively rather than supplied by the user. We focus on broad, recurring behaviors rather than task-specific corner cases, and provide concrete examples in the case studies in App.~\ref{app:case_study}.

\paragraph{Ignoring recoverable prior context.}
Some agents treat the current user message as a standalone request even when the task depends on information established in earlier sessions. This leads to missed hidden intents or incorrect final artifacts, as shown in the Researcher dependency failure in Fig.~\ref{fig:case_leihaodi_2_kimi_trajectory} and the meal-planning contrast in Fig.~\ref{fig:case_zhangshunkai_1_deepseek_trajectory}.

\paragraph{Completing the visible request while missing hidden requirements.}
Agents often produce a plausible answer for the explicit request but leave implicit preferences, formatting constraints, or task-specific requirements unresolved until the user states them. Fig.~\ref{fig:case_zhanghaoran_1_hidden_intent_resolution} illustrates this pattern through several hidden intents marked as user-provided rather than agent-completed, and Fig.~\ref{fig:case_zhangshunkai_1_claude_trajectory} shows that even a response with high checklist completeness can still rely on user-provided hidden requirements.

\paragraph{Failing to ask targeted clarifications.}
When information is genuinely missing, clarification is useful only if it targets the specific requirement needed to complete the task. The targeted-followup successes in Fig.~\ref{fig:case_zhangshunkai_2_targeted_followup_trajectory} and Fig.~\ref{fig:case_zhangshunkai_3_targeted_followup_trajectory} serve as positive contrasts: weaker trajectories instead rely on repeated user intervention before hidden requirements are surfaced.

\paragraph{Using tools without verifying the required artifact.}
Some failures occur after the agent invokes relevant tools but does not verify that the produced artifact contains the required content or that the final state matches the checklist. The Law Trainee comparison in Fig.~\ref{fig:case_leihaodi_1_kimi_trajectory} shows this gap: the agent uses SMS and Todoist tools, but the final trace does not preserve the required handover details or reminder checks.

\section{Task Format}
\label{app:task_format}

Each task is stored as a structured configuration that specifies the user-facing request, latent requirements, evaluation objectives, and metadata. The example below shows a survey-paper task, where earlier tasks in the same group may give the agent opportunities to infer the hidden intents needed for this final task. Task-specific names, links, and sensitive details are anonymized.

\begin{tcbCodeListing}{Example Task Format}{tcbyaml}
title: "Survey Recent Papers on a User-Specified Research Topic"

description: >
  The user is starting a small research project and asks the agent to organize
  a set of relevant papers from a local paper list. The task requires the agent
  to infer the user's preferred topic focus, filter candidate papers, and
  prepare a concise reading plan that supports later topic selection.

task_type: long_term

trigger:
  type: user

intent:
  initial_input: >
    I wrote a local file at [FILE_NAME] that contains recently accepted papers.
    Please organize several papers related to my research theme for me.

  hidden_intent:
    - content: "Prioritize papers related to [RESEARCH_THEME]."
    - content: "Recommend around [N] papers rather than producing a long survey."
    - content: >
        For each paper, provide a brief introduction and summarize the key
        technical points.
    - content: >
        Include the official paper link when available.
    - content: >
        Include the project or code link when available; otherwise explicitly
        state that no public repository is found.
    - content: >
        For each paper, state whether it is suitable to follow and explain how
        the user should follow it.

objectives:
  checklist:
    - criterion: "Include [PAPER_TITLE_1]."
    - criterion: >
        Explain that [PAPER_TITLE_1] studies [CORE_METHOD_1] for
        [RESEARCH_THEME], and mention its main training or inference strategy.
    - criterion: "Write the official paper link for [PAPER_TITLE_1] as [PAPER_LINK_1]."
    - criterion: "Write the code link for [PAPER_TITLE_1] as [CODE_LINK_1]."

    - ......

    - criterion: >
        For every recommended paper, state whether it is suitable to follow.
    - criterion: >
        For every recommended paper, give a concrete follow-up action, such as
        reading the method section, reproducing the code, or comparing the task
        formulation with the user's project.

metadata:
  difficulty: hard
\end{tcbCodeListing}

\section{Discussion}

\subsection{Metric Relationship}
\label{app:metric_relationship}
Proactivity and completeness are related but intentionally separated. In real user-agent interaction, a final outcome may improve either because the agent proactively identifies hidden requirements or because the user eventually spells them out. {\bench} follows the latter interaction to exhaustion. The simulated user continues until every hidden intent has been completed, elicited, or explicitly provided. Thus, the final trajectory contains the complete set of user requirements, either through the agent's own initiative or through oracle-like user provision. Under this protocol, \textsc{Proc} measures who drives requirement discovery, while \textsc{Comp} measures whether the agent converts the resulting trajectory into correct artifacts and decisions. The two scores therefore need not move together. A reactive agent can obtain high \textsc{Comp} after receiving many user-provided intents, while a proactive agent can still lose \textsc{Comp} by mishandling files, tools, formats, or downstream constraints. This separation lets us analyze the trade-off between surfacing latent requirements and completing the final workflow.

\subsection{Hidden Intents and Checklists}
\label{app:hidden_intents_checklists}

Hidden intents and checklists serve different roles in our benchmark.

\paragraph{Hidden intents.} Hidden intents describe latent requirements that are not stated in the initial request but should guide the agent's behavior during interaction. They capture what a proactive assistant should infer, ask about, or act on before the user explicitly provides the information. Such intents may involve user preferences, task constraints, output conventions, or cross-session dependencies. Therefore, hidden intents are used to evaluate how the agent handles underspecification and whether it reduces the need for user intervention.

\paragraph{Checklists.}
Checklists, in contrast, define verifiable criteria for task completion. They specify what must be true of the final trajectory or produced artifacts, such as whether a required file is created, whether an output follows the requested format, whether a tool action succeeds, or whether a generated artifact contains the necessary content. A checklist item may depend on a hidden intent, but it is evaluated only as an outcome requirement. This separation lets us distinguish proactive behavior from final correctness. An agent may complete a task after the user provides all missing requirements, leading to high completeness but lower proactivity. Conversely, an agent may correctly identify hidden intents but still fail to satisfy some concrete checklist items, leading to high proactivity but lower completeness.
\section{Benchmark Construction and Annotation}
\label{app:data_construction}

\subsection{Task Sources and Workflow Collection}
\label{app:task_sources}

\paragraph{Workflow sources.}

Tasks are derived from realistic workflows associated with the five user roles in {\bench}: researcher, marketer, law trainee, pharmacist, and financier. For each role, we first collect representative work routines, common deliverables, and supporting materials from domain experts and public task patterns. These sources cover both professional work and everyday knowledge work, such as experiment analysis, paper writing, content planning, document review, literature organization, financial reconciliation, and report generation.

\paragraph{Selection criteria.}
During collection, we focus on workflows that are artifact-centric and naturally require interaction. We also require the workflow to contain realistic underspecification. In other words, the task should not be fully determined by the first user message. Instead, correct completion should depend on constraints, preferences, prior decisions, or workspace context that the agent needs to recover through proactive behavior.

\paragraph{Data sanitization.}
We do not directly use private or sensitive real-world data. When a workflow is inspired by actual work practice, annotators rewrite and normalize the materials into synthetic but realistic task instances. This includes replacing private names, removing sensitive details, simplifying irrelevant background, and ensuring that all required information can be accessed through the provided workspace, memory, or interaction protocol.

\subsection{Task Construction Procedure}
\label{app:task_construction_procedure}

\paragraph{Task goal selection.}
For each candidate workflow, annotators first define a task goal that is realistic, checkable, and well-scoped. The goal should reflect a plausible need of the corresponding user role, cover a meaningful part of the workflow, and admit clear evidence for whether the agent has completed it. We avoid goals that are either too narrow to require interaction or too broad to support reliable evaluation.

\paragraph{Task specification.}
Annotators then convert the workflow into a concrete task specification. This process includes:
\begin{itemize}[leftmargin=10pt, topsep=0pt, itemsep=1pt, partopsep=1pt, parsep=1pt]
    \item Decide what information should appear in the initial request and what information should remain implicit as hidden intents.
    \item Prepare the workspace state and supporting materials, such as input files, prior artifacts, structured records, or tool-accessible information.
    \item Specify cross-session dependencies when earlier decisions, files, or user preferences should influence the current session.
\end{itemize}
The initial request must be underspecified enough to test proactivity, while still being concrete enough for the agent to begin useful work.

\paragraph{Intent and checklist annotation.}
Annotators next define the hidden intents and checklist items. Hidden intents capture latent requirements that should affect the agent's behavior during the interaction, such as constraints, preferences, output conventions, or dependencies on prior context. Checklist items capture final completion requirements that can be verified from the trajectory, produced artifacts, or tool records.

\paragraph{Feasibility validation.}
After annotation, annotators validate each task before inclusion in the benchmark. This step checks whether the task is solvable with the provided context, whether the required tools and files are correctly connected, and whether the evaluation criteria can be applied reliably. The validation process includes:
\begin{itemize}[leftmargin=10pt, topsep=0pt, itemsep=1pt, partopsep=1pt, parsep=1pt]
    \item Construct a reference workflow that describes a plausible completion path using the available workspace, tools, and interaction protocol.
    \item Create expected artifacts when the task requires concrete outputs, such as documents, tables, code files, or structured records.
    \item Bind required tools, input files, and workspace paths to ensure that the agent can access the information needed for completion.
    \item Run pilot executions to identify missing context, broken tool paths, unclear instructions, unstable graders, or checklist items that are difficult to judge.
\end{itemize}
Tasks are revised or removed when validation reveals ambiguity or infeasibility. Revisions may include rewriting the initial request, refining hidden intents, splitting checklist items, adding missing workspace context, or adjusting graders for more stable verification.

\subsection{Annotation Guidelines}
\label{app:annotation_guidelines}

\paragraph{Initial requests.}
The initial request defines the user-facing starting point of a session. Annotators write it to be natural and actionable, while deliberately leaving some task-relevant requirements unstated so that proactive behavior can be evaluated. Tab.~\ref{tab:annotation_initial_request} summarizes the main annotation rules.

\begin{table}[tb]
\centering
\caption{Annotation guidelines for initial requests.}
\label{tab:annotation_initial_request}
\small
\begin{tabular}{p{0.20\linewidth}p{0.36\linewidth}p{0.34\linewidth}}
\toprule[1.25pt]
\textbf{Aspect} & \textbf{Do} $\checkmark$ & \textbf{Avoid} $\times$ \\
\midrule[0.75pt]
\textsc{Naturalness} &
Use concise wording that matches the user's role and working style. &
Do not use artificial benchmark-like phrasing or expose internal task metadata. \\
\midrule[0.75pt]
\textsc{Actionability} &
State a clear surface goal so that the agent can begin useful work. &
Do not make the request so vague that the agent must ask the user to restate the task. \\
\midrule[0.75pt]
\textsc{Underspecification} &
Omit at least one task-relevant requirement that the agent should infer, ask about, or recover from context. &
Do not include all constraints, preferences, and output requirements in the initial request. \\
\midrule[0.75pt]
\textsc{Environment fit} &
Make the request consistent with the persona, workspace state, and available tool interfaces when applicable. &
Do not refer to files, records, or actions that are unavailable in the task environment. \\
\bottomrule[1.25pt]
\end{tabular}
\end{table}

\paragraph{Hidden intents.}
Hidden intents specify unstated requirements that should shape the agent's behavior during interaction. Annotators use them to define what a proactive assistant should complete directly or elicit through a targeted question. Tab.~\ref{tab:annotation_hidden_intents} summarizes the main rules.

\begin{table}[tb]
\centering
\caption{Annotation guidelines for hidden intents.}
\label{tab:annotation_hidden_intents}
\small
\begin{tabular}{p{0.20\linewidth}p{0.36\linewidth}p{0.34\linewidth}}
\toprule[1.25pt]
\textbf{Aspect} & \textbf{Do} $\checkmark$ & \textbf{Avoid} $\times$ \\
\midrule[0.75pt]
\textsc{Unstated requirement} &
Each hidden intent must be absent from the initial request. &
Do not restate information that is already explicitly given to the agent. \\
\midrule[0.75pt]
\textsc{Task relevance} &
Each hidden intent should affect task handling, output quality, or downstream decisions. &
Do not add arbitrary preferences that have no effect on the task. \\
\midrule[0.75pt]
\textsc{Specificity} &
Write each intent precisely enough to support terminal status assignment during interaction. &
Do not use vague intents such as ``make it better'' or ``follow best practice'' without a concrete meaning. \\
\midrule[0.75pt]
\textsc{Scope} &
Mark whether the intent is session-local or persistent across sessions when relevant. &
Do not let persistent intents appear without supporting evidence in prior context or workspace state. \\
\bottomrule[1.25pt]
\end{tabular}
\end{table}

\paragraph{Checklists and graders.}
Checklist items define verifiable outcome criteria for completeness evaluation. Annotators write them separately from hidden intents so that proactive intent resolution and final task completion can be measured independently. Tab.~\ref{tab:annotation_checklists} summarizes the main rules.

\begin{table}[tb]
\centering
\caption{Annotation guidelines for checklist items and grader assignment.}
\label{tab:annotation_checklists}
\small
\begin{tabular}{p{0.18\linewidth}p{0.37\linewidth}p{0.35\linewidth}}
\toprule[1.25pt]
\textbf{Aspect} & \textbf{Do} $\checkmark$ & \textbf{Avoid} $\times$ \\
\midrule[0.75pt]
\textsc{Verifiability} &
Each item should be checkable from the trajectory, produced artifacts, or tool records. &
Do not include criteria that require unstated assumptions or external judgment beyond the available evidence. \\
\midrule[0.75pt]
\textsc{Atomicity} &
Each item should cover one requirement whenever possible. &
Do not combine several independent requirements into one checklist item. \\
\midrule[0.75pt]
\textsc{Outcome focus} &
Write checklist items as final completion conditions. &
Do not use checklist items to score whether the agent was proactive during the interaction. \\
\midrule[0.75pt]
\textsc{Grader choice} &
Use rubric-based evaluation for semantic criteria and rule-based verification for exact structured criteria. &
Do not use an LLM rubric when an exact programmatic check is available and reliable. \\
\midrule[0.75pt]
\textsc{Auditability} &
Make the expected evidence clear enough for another expert to inspect the judgment. &
Do not write criteria whose satisfaction cannot be traced to concrete evidence. \\
\bottomrule[1.25pt]
\end{tabular}
\end{table}

\subsection{Quality Control}
\label{app:data_quality_control}

Before inclusion in the final benchmark, each task undergoes iterative checks and revisions. Annotators inspect the task package for leakage and recoverability, ensuring that hidden intents are not directly exposed in the initial request while still being recoverable from prior sessions, workspace artifacts, memory, tool results, or targeted clarification. They also run pilot executions to detect setup issues, including missing files, invalid tool bindings, broken workspace paths, and infeasible task flows. Finally, they validate the evaluation resources by checking that rubric-based criteria can be judged from the selected evidence and that rule-based graders behave correctly on expected cases. Tasks are revised or removed when these checks reveal ambiguity, infeasibility, or unstable evaluation.
\clearpage
\section{Case Study}
\label{app:case_study}

\subsection{Hidden Intents}

\paragraph{Failure case: Researcher -- Claude 4.6 Opus.}
Fig.~\ref{fig:case_zhanghaoran_1_hidden_intent_trigger} and Fig.~\ref{fig:case_zhanghaoran_1_hidden_intent_resolution} show a representative hidden-intent case from the Researcher user episode evaluated with Claude 4.6 Opus.
The session begins from an environment-triggered paper recommendation feed, where the relevant requirement is not stated as a direct user instruction.
The agent first handles the feed as a broad recommendation task, and only later focuses on the OpenClaw-related subset after the user explicitly asks for it.
The key signal is the hidden-intent assignment: several intents are marked as {\provided} rather than {\completed} or {\inferred}, indicating that the user had to surface requirements that the agent did not proactively resolve.
Although the final trajectory has moderate completeness, this case mainly illustrates weak proactive intent resolution under an underspecified trigger.

\providecommand{\caseZhanghaoranOneTag}[3]{%
    {\setlength{\fboxsep}{1.6pt}\colorbox{#1}{\textcolor{#2}{\scriptsize\textbf{#3}}}}%
}
\providecommand{\caseZhanghaoranOneLine}[3]{%
    \caseZhanghaoranOneTag{#1}{#2}{#3}\hspace{3pt}%
}
\providecommand{\caseZhanghaoranOneIO}[3]{%
    \vspace{2pt}
    \noindent\caseZhanghaoranOneLine{#1}{white}{#2}\textbf{#3}
}
\providecommand{\caseZhanghaoranOneNote}[1]{%
    \vspace{2pt}
    \noindent{\scriptsize\color{black!55}\textit{#1}}
}

\begin{figure}[h]
    \centering
    \tcolorboxCase{Case Study -- Researcher -- Claude 4.6 Opus: Task and Trigger}{
    \small
    \caseZhanghaoranOneLine{purple!70!black}{white}{$\blacktriangleright$ TASK} Filter papers the user cares about most from a simulated Hugging Face daily-papers feed.
    The session is started by an environment event, so the agent must decide whether the event is actionable without waiting for a direct user command.

    \vspace{3pt}
    \caseZhanghaoranOneLine{blue!65!black}{white}{$\rightarrow$ INITIAL REQUEST}
    \caseZhanghaoranOneNote{This is an environment-triggered initial request. The event lists candidate papers, but does not explicitly state the OpenClaw filter or the link and metadata requirements.}
    \begin{quote}
    \tiny
    \setlength{\baselineskip}{6.2pt}
    \raggedright
    \texttt{source=huggingface\_hub; event=paper\_recommendation\_trigger; target=agent; papers=}\par
    \texttt{paper\_01}: Common Corpus: The Largest Collection of Ethical Data for LLM Pre-Training;\par
    \texttt{paper\_02}: Q-RAG: Long Context Multi-Step Retrieval via Value-Based Embedder Training;\par
    \texttt{paper\_03}: From movement to cognitive maps: recurrent neural networks reveal how locomotor development shapes hippocampal spatial coding;\par
    \texttt{paper\_04}: FIRE: Frobenius-Isometry Reinitialization for Balancing the Stability-Plasticity Tradeoff;\par
    \texttt{paper\_05}: Exchangeability of GNN Representations with Applications to Graph Retrieval;\par
    \texttt{paper\_06}: A Trajectory-Based Safety Audit of Clawdbot (OpenClaw);\par
    \texttt{paper\_07}: Enhancing Generative Auto-bidding with Offline Reward Evaluation and Policy Search;\par
    \texttt{paper\_08}: Why DPO is a Misspecified Estimator and How to Fix It;\par
    \texttt{paper\_09}: WebDevJudge: Evaluating (M)LLMs as Critiques for Web Development Quality;\par
    \texttt{paper\_10}: SafeDPO: A Simple Approach to Direct Preference Optimization with Enhanced Safety;\par
    \texttt{paper\_11}: MedAgentGym: A Scalable Agentic Training Environment for Code-Centric Reasoning in Biomedical Data Science;\par
    \texttt{paper\_12}: Optimistic Task Inference for Behavior Foundation Models;\par
    \texttt{paper\_13}: CounselBench: A Large-Scale Expert Evaluation and Adversarial Benchmarking of Large Language Models in Mental Health Question Answering;\par
    \texttt{paper\_14}: AstaBench: Rigorous Benchmarking of AI Agents with a Scientific Research Suite;\par
    \texttt{paper\_15}: MetaClaw: Just Talk An Agent That Meta-Learns and Evolves in the Wild;\par
    \texttt{paper\_16}: Neon: Negative Extrapolation From Self-Training Improves Image Generation;\par
    \texttt{paper\_17}: Compositional Diffusion with Guided search for Long-Horizon Planning;\par
    \texttt{paper\_18}: Visual symbolic mechanisms: Emergent symbol processing in Vision Language Models;\par
    \texttt{paper\_19}: Addressing divergent representations from causal interventions on neural networks;\par
    \texttt{paper\_20}: Cross-Domain Lossy Compression via Rate- and Classification-Constrained Optimal Transport;\par
    \texttt{paper\_21}: Latent Fourier Transform;\par
    \texttt{paper\_22}: GLASS Flows: Efficient Inference for Reward Alignment of Flow and Diffusion Models;\par
    \texttt{paper\_23}: Latent Speech-Text Transformer;\par
    \texttt{paper\_24}: LoongRL: Reinforcement Learning for Advanced Reasoning over Long Contexts;\par
    \texttt{paper\_25}: Improving Developer Emotion Classification via LLM-Based Augmentation;\par
    \texttt{paper\_26}: Revisiting Multilingual Data Mixtures in Language Model Pretraining;\par
    \texttt{paper\_27}: One-Shot Style Personalization for RL Agents via Latent Discriminator;\par
    \texttt{paper\_28}: OpenClaw-RL: Train Any Agent Simply by Talking;\par
    \texttt{paper\_29}: Compositional HyperModules for Few-Shot Code Adaptation in Meta-Reinforcement Learning;\par
    \texttt{paper\_30}: All-in-One: Boosting Basic Capabilities in one Omni-MLLM to Enhance Movie Understanding;\par
    \texttt{paper\_31}: Contrastive Code Graph Embeddings for Reinforcement Learning-Based Automated Code Refactoring;\par
    \texttt{paper\_32}: Soft Non-Diagonality Penalty Enables Latent Space-Level Interpretability of pLM at No Performance Cost;\par
    \texttt{paper\_33}: Adaptive Mixing of Non-Invariant Information for Generalized Diffusion Policy;\par
    \texttt{paper\_34}: Style2Shape: Image Style Guided 3D Shape Material Generation;\par
    \texttt{paper\_35}: Contrastive-Aligned Knowledge Distillation for Collaborative Code Completion via Multi-Agent Reinforcement Learning;\par
    \texttt{paper\_36}: Scaling Laws for Generative Reward Models;\par
    \texttt{paper\_37}: Contrastive-Online-Meta (COM): A Dynamic Adaptation Mechanism for Instruction-Tuned CodeLLMs;\par
    \texttt{paper\_38}: FedPAC: Consistent Representation Learning for Federated Unsupervised Learning under Data Heterogeneity;\par
    \texttt{paper\_39}: Training as Computation: A Resource-Bounded Theory of Continual Self-Play Learning;\par
    \texttt{paper\_40}: Cross-Modal Syntax-NL Attention for Multi-Agent Reinforcement Learning in Collaborative Coding;\par
    \texttt{paper\_41}: Less is More: Improving Molecular Force Fields with Minimal Temporal Information;\par
    \texttt{paper\_42}: Curricular Adversarial Training for Robust Code Generation via Hierarchical Reinforcement Learning;\par
    \texttt{paper\_43}: ``Humans welcome to observe'': A First Look at the Agent Social Network Moltbook;\par
    \texttt{paper\_44}: Triple-S: A Sticker Semantic Similarity Benchmark with General Sticker Encoder;\par
    \texttt{paper\_45}: Unsupervised Dynamic Graph Multi-Model Representation Learning for Temporal Patterns Discovery: Uncovering Parkinson's Disease Stages Using Cerebrospinal Fluid Longitudinal Profiles;\par
    \texttt{paper\_46}: Causal Reasoning Favors Encoders: Limits of Decoder-Only Models;\par
    \texttt{paper\_47}: Hierarchies over Pixels: A Benchmark for Cognitive Geospatial Reasoning for Agents;\par
    \texttt{paper\_48}: But what is your honest answer? Aiding LLM-judges with honest alternatives using steering vectors;\par
    \texttt{paper\_49}: LaMbDA: Local Latent Embedding Alignment for Cross-modal Time-Series Diffusion.
    \end{quote}

    \vspace{2pt}
    \caseZhanghaoranOneLine{orange!80!black}{white}{$\star$ HIDDEN INTENTS}
    \par
    \caseZhanghaoranOneTag{red!12}{red!70!black}{$\circ$ I1} List the OpenClaw-related papers from the recommendation trigger.
    \par
    \caseZhanghaoranOneTag{red!12}{red!70!black}{$\circ$ I2} Add a short introduction and key technical points for each OpenClaw-related paper.
    \par
    \caseZhanghaoranOneTag{red!12}{red!70!black}{$\circ$ I3} Include arXiv links for the OpenClaw-related papers.
    \par
    \caseZhanghaoranOneTag{red!12}{red!70!black}{$\circ$ I4} Include GitHub links for the OpenClaw-related papers.
    \par
    \caseZhanghaoranOneTag{red!12}{red!70!black}{$\circ$ I5} Give the research institution and authors for each OpenClaw-related paper.
    }
    \caption{Task trigger and hidden intents.}
    \label{fig:case_zhanghaoran_1_hidden_intent_trigger}
\end{figure}

\begin{figure}[h]
    \centering
    \tcolorboxCase{Case Study -- Researcher -- Claude 4.6 Opus: Trajectory and Scores}{
    \small
    \caseZhanghaoranOneIO{blue!65!black}{$\rightarrow$ INPUT}{Turn 1.}
    Environment trigger with the 49-paper payload in Fig.~\ref{fig:case_zhanghaoran_1_hidden_intent_trigger}.

    \caseZhanghaoranOneIO{green!45!black}{$\Leftarrow$ OUTPUT SUMMARY}{Turn 1.}
    The agent filters the batch by broad interests: ``think-with-image,'' multimodal reasoning, multimodal tool use, and VLM/MLLM agents. It highlights several adjacent multimodal papers, but does not surface the annotated OpenClaw-related subset.

    \vspace{3pt}
    \caseZhanghaoranOneIO{blue!65!black}{$\rightarrow$ INPUT}{Turn 2.}
    \emph{Give list OpenClaw-related papers from this recommendation trigger.}
    \caseZhanghaoranOneTag{red!15}{red!70!black}{provided: I1}

    \caseZhanghaoranOneIO{green!45!black}{$\Leftarrow$ OUTPUT}{Turn 2.}
    \emph{From the recommendation batch, here are the OpenClaw-related papers: A Trajectory-Based Safety Audit of Clawdbot (OpenClaw); MetaClaw: Just Talk An Agent That Meta-Learns and Evolves in the Wild; OpenClaw-RL: Train Any Agent Simply by Talking. That's 3 papers referencing the OpenClaw ecosystem.}

    \vspace{3pt}
    \caseZhanghaoranOneIO{blue!65!black}{$\rightarrow$ INPUT}{Turn 3.}
    \emph{Give add a short introduction and key technical points for each OpenClaw-related paper.}
    \caseZhanghaoranOneTag{red!15}{red!70!black}{provided: I2}

    \caseZhanghaoranOneIO{green!45!black}{$\Leftarrow$ OUTPUT SUMMARY}{Turn 3.}
    The agent searches web sources, fetches arXiv pages and a project page, and writes introductions plus technical points. It also includes several arXiv and GitHub links. \caseZhanghaoranOneTag{green!12}{green!45!black}{completed: I3}\hspace{2pt}\caseZhanghaoranOneTag{green!12}{green!45!black}{completed: I4}

    \vspace{3pt}
    \caseZhanghaoranOneIO{blue!65!black}{$\rightarrow$ INPUT}{Turn 4.}
    \emph{Give the research institution and author for each OpenClaw-related paper.}
    \caseZhanghaoranOneTag{red!15}{red!70!black}{provided: I5}

    \caseZhanghaoranOneIO{green!45!black}{$\Leftarrow$ OUTPUT SUMMARY}{Turn 4.}
    The agent searches for author and affiliation metadata and then reports institutions and authors for the three selected papers.

    \vspace{5pt}
    \caseZhanghaoranOneLine{purple!70!black}{white}{$\blacktriangleright$ EVALUATION}
    \caseZhanghaoranOneTag{orange!15}{orange!80!black}{\textsc{Comp}=65.00}
    \hspace{3pt}
    \caseZhanghaoranOneTag{cyan!15}{cyan!45!black}{\textsc{Proc}=40.00}

    \vspace{3pt}
    \noindent\caseZhanghaoranOneLine{orange!80!black}{white}{$\star$ CHECKLIST SIGNALS}
    Satisfied: several paper recommendations, selected links, and selected author/institution fields.
    Not satisfied: full annotated paper coverage and some detailed metadata criteria.
    }
    \caption{Trajectory and scores; colored intent tags indicate terminal status, e.g., \caseZhanghaoranOneTag{red!15}{red!70!black}{provided: I1} denotes a user-provided intent and \caseZhanghaoranOneTag{green!12}{green!45!black}{completed: I3} denotes an agent-completed intent.}
    \label{fig:case_zhanghaoran_1_hidden_intent_resolution}
\end{figure}

\paragraph{Contrast case: Researcher -- DeepSeek V3.2 and Claude 4.6 Opus.}
Fig.~\ref{fig:case_zhangshunkai_1_deepseek_task} and Fig.~\ref{fig:case_zhangshunkai_1_deepseek_trajectory} show the DeepSeek V3.2 trace, while Fig.~\ref{fig:case_zhangshunkai_1_claude_task} and Fig.~\ref{fig:case_zhangshunkai_1_claude_trajectory} show the Claude 4.6 Opus trace for the same Researcher meal-planning task.
Both runs receive the same initial request, which explicitly mentions only a one-week meal plan and the RMB 20--30 price constraint, while the user's body profile, macro accounting, table structure, and Plan B requirements remain hidden but recoverable from prior context.
We focus on the first assistant turn because it occurs before the user reveals any hidden requirements and therefore best isolates proactive recovery from visible-request following.
DeepSeek V3.2 mostly follows the visible budget and availability cue in this first response; although it recognizes campus-accessible food options, it does not recover the prior-session body profile, table preference, per-meal macro requirement, or Plan B preference, so the user has to reveal several requirements across later turns.
Claude 4.6 Opus captures more task structure immediately: it goes beyond the visible price constraint by imposing a readable table, listing concrete foods for each day and meal, and grounding choices in campus or nearby meal availability.
However, Claude still requires later user turns for macro totals, per-meal P/F/C values, and fallback meals.
The contrast is a stronger-versus-weaker proactivity comparison rather than a binary success-failure case: both trajectories improve after user intervention, but their scores differ according to how much hidden intent is resolved before the user supplies it.

\providecommand{\caseZhangshunkaiOneTag}[3]{%
    {\setlength{\fboxsep}{1.6pt}\colorbox{#1}{\textcolor{#2}{\scriptsize\textbf{#3}}}}%
}
\providecommand{\caseZhangshunkaiOneLine}[3]{%
    \caseZhangshunkaiOneTag{#1}{#2}{#3}\hspace{3pt}%
}
\providecommand{\caseZhangshunkaiOneIO}[3]{%
    \vspace{2pt}
    \noindent\caseZhangshunkaiOneLine{#1}{white}{#2}\textbf{#3}
}
\providecommand{\caseZhangshunkaiOneNote}[1]{%
    \vspace{2pt}
    \noindent{\scriptsize\color{black!55}\textit{#1}}
}

\begin{figure}[h]
    \centering
    \tcolorboxCase{Case Study -- Researcher -- DeepSeek V3.2: Task and Hidden Intents}{
    \small
    \caseZhangshunkaiOneLine{purple!70!black}{white}{$\blacktriangleright$ TASK}
    Design a one-week meal plan with per-meal prices controlled within RMB 20--30.
    The visible request only asks for a meal plan and prices, while the full task also requires body-profile conditioning, macro accounting, table structure, obtainable foods, and fallback meals.

    \vspace{3pt}
    \caseZhangshunkaiOneLine{purple!70!black}{white}{$\blacktriangleright$ PRIOR CONTEXT}
    In a prior muscle-gain gear-selection session, the user had already exposed the same body profile, the muscle-gain goal, and preferences for readable tables and Plan B alternatives.

    \vspace{3pt}
    \caseZhangshunkaiOneLine{blue!65!black}{white}{$\rightarrow$ INITIAL REQUEST}
    \begin{quote}
    \small
    \raggedright
    Help me design a one-week meal plan. Provide the price of each meal, controlled within RMB 20--30.
    \end{quote}

    \vspace{2pt}
    \caseZhangshunkaiOneLine{orange!80!black}{white}{$\star$ PARTIAL HIDDEN INTENTS}
    \par
    \caseZhangshunkaiOneTag{red!12}{red!70!black}{$\circ$ I1} Use the user's body profile: about 175 cm, 68 kg, 17\% body fat, and muscle-gain phase.
    \par
    \caseZhangshunkaiOneTag{red!12}{red!70!black}{$\circ$ I2} Provide daily totals for protein, fat, and carbohydrates in grams.
    \par
    \caseZhangshunkaiOneTag{red!12}{red!70!black}{$\circ$ I3} Use a readable table format.
    \par
    \caseZhangshunkaiOneTag{red!12}{red!70!black}{$\circ$ I4} List specific foods for each meal on each day.
    \par
    \caseZhangshunkaiOneTag{red!12}{red!70!black}{$\circ$ I5} Prefer meals obtainable from school cafeterias, nearby shops, or common takeout platforms.
    \par
    \caseZhangshunkaiOneTag{red!12}{red!70!black}{$\circ$ I6} Give per-meal protein, fat, and carbohydrate amounts in grams.
    \par
    \caseZhangshunkaiOneTag{red!12}{red!70!black}{$\circ$ I7} Provide a substitute option, Plan B, for each meal.
    }
    \caption{Task trigger and partial hidden intents.}
    \label{fig:case_zhangshunkai_1_deepseek_task}
\end{figure}

\begin{figure}[h]
    \centering
    \tcolorboxCase{Case Study -- Researcher -- DeepSeek V3.2: Trajectory and Scores}{
    \small
    \caseZhangshunkaiOneIO{blue!65!black}{$\rightarrow$ INPUT}{Turn 1.}
    Initial request in Fig.~\ref{fig:case_zhangshunkai_1_deepseek_task}.

    \caseZhangshunkaiOneIO{green!45!black}{$\Leftarrow$ OUTPUT SUMMARY}{Turn 1.}
    The agent creates a loose day-by-day meal plan with budget notes, Fudan/campus pricing assumptions, and availability-oriented cost-saving tips.
    It does not yet use the body profile, daily macro targets, per-meal macro values, a compact readable meal-plan table, or per-meal Plan B substitutes.
    \caseZhangshunkaiOneTag{green!12}{green!45!black}{completed: I5}

    \vspace{3pt}
    \caseZhangshunkaiOneIO{blue!65!black}{$\rightarrow$ INPUT}{Turn 2.}
    \emph{Height about 175 cm, weight about 68 kg, body-fat ratio about 17\%; currently in a muscle-gain phase.}
    \caseZhangshunkaiOneTag{red!15}{red!70!black}{provided: I1}

    \caseZhangshunkaiOneIO{green!45!black}{$\Leftarrow$ OUTPUT SUMMARY}{Turn 2.}
    The agent revises the plan for muscle gain, mentions higher protein intake and calorie surplus, and reports rough protein totals, but still does not produce the required daily macro totals, per-meal macro table, or Plan B structure.

    \vspace{3pt}
    \caseZhangshunkaiOneIO{blue!65!black}{$\rightarrow$ INPUT}{Turn 3.}
    \emph{Based on the user's body weight, provide the required daily totals of the three macronutrients (protein/fat/carbohydrates) for each day of the week, in grams (g).}
    \caseZhangshunkaiOneTag{red!15}{red!70!black}{provided: I2}

    \caseZhangshunkaiOneIO{green!45!black}{$\Leftarrow$ OUTPUT}{Turn 3.}
    \emph{Based on your body weight (68kg) and muscle-gain goals, I'll calculate the daily macronutrient requirements for each day of the week.}
    The response then gives protein, carbohydrate, and fat targets, distinguishing training from rest days.

    \vspace{3pt}
    \caseZhangshunkaiOneIO{blue!65!black}{$\rightarrow$ INPUT}{Turn 4.}
    \emph{For each meal, indicate macronutrient amounts (protein/fat/carbohydrates), in grams (g).}
    \caseZhangshunkaiOneTag{red!15}{red!70!black}{provided: I6}

    \caseZhangshunkaiOneIO{green!45!black}{$\Leftarrow$ OUTPUT SUMMARY}{Turn 4.}
    The agent provides detailed per-meal macro breakdowns for the revised plan, but the response remains verbose and does not yet include the required Plan B substitutes.

    \vspace{3pt}
    \caseZhangshunkaiOneIO{blue!65!black}{$\rightarrow$ INPUT}{Turn 5.}
    \emph{Provide a substitute option (Plan B) for each meal to handle unexpected situations.}
    \caseZhangshunkaiOneTag{red!15}{red!70!black}{provided: I7}

    \caseZhangshunkaiOneIO{green!45!black}{$\Leftarrow$ OUTPUT}{Turn 5.}
    \emph{I'll provide substitute options (Plan B) for each meal to handle unexpected situations like campus canteen closures, food availability issues, or budget constraints.}
    The response then adds substitute meals and emergency backup options after the user explicitly asks for Plan B.

    \vspace{5pt}
    \caseZhangshunkaiOneLine{purple!70!black}{white}{$\blacktriangleright$ EVALUATION}
    \caseZhangshunkaiOneTag{orange!15}{orange!80!black}{\textsc{Comp}=87.50}
    \hspace{3pt}
    \caseZhangshunkaiOneTag{cyan!15}{cyan!45!black}{\textsc{Proc}=42.86}

    \vspace{3pt}
    \noindent\caseZhangshunkaiOneLine{orange!80!black}{white}{$\star$ CHECKLIST SIGNALS}
    Final satisfied: C1, C2, C3, C4, C6, C7, C8.
    Not satisfied: C5, the final table does not reliably list the specific foods for each meal each day.
    }
    \caption{DeepSeek trajectory and scores.}
    \label{fig:case_zhangshunkai_1_deepseek_trajectory}
\end{figure}

\begin{figure}[h]
    \centering
    \tcolorboxCase{Case Study -- Researcher -- Claude 4.6 Opus: Task and Hidden Intents}{
    \small
    \caseZhangshunkaiOneLine{purple!70!black}{white}{$\blacktriangleright$ TASK}
    Design a one-week meal plan with per-meal prices controlled within RMB 20--30.
    The visible request only asks for a meal plan and prices, while the full task also requires body-profile conditioning, macro accounting, table structure, obtainable foods, and fallback meals.

    \vspace{3pt}
    \caseZhangshunkaiOneLine{purple!70!black}{white}{$\blacktriangleright$ PRIOR CONTEXT}
    In a prior muscle-gain gear-selection session, the user had already exposed the same body profile, the muscle-gain goal, and preferences for readable tables and Plan B alternatives.

    \vspace{3pt}
    \caseZhangshunkaiOneLine{blue!65!black}{white}{$\rightarrow$ INITIAL REQUEST}
    \begin{quote}
    \small
    \raggedright
    Help me design a one-week meal plan. Provide the price of each meal, controlled within RMB 20--30.
    \end{quote}

    \vspace{2pt}
    \caseZhangshunkaiOneLine{orange!80!black}{white}{$\star$ PARTIAL HIDDEN INTENTS}
    \par
    \caseZhangshunkaiOneTag{red!12}{red!70!black}{$\circ$ I1} Use the user's body profile: about 175 cm, 68 kg, 17\% body fat, and muscle-gain phase.
    \par
    \caseZhangshunkaiOneTag{red!12}{red!70!black}{$\circ$ I2} Provide daily totals for protein, fat, and carbohydrates in grams.
    \par
    \caseZhangshunkaiOneTag{red!12}{red!70!black}{$\circ$ I3} Use a readable table format.
    \par
    \caseZhangshunkaiOneTag{red!12}{red!70!black}{$\circ$ I4} List specific foods for each meal on each day.
    \par
    \caseZhangshunkaiOneTag{red!12}{red!70!black}{$\circ$ I5} Prefer meals obtainable from school cafeterias, nearby shops, or common takeout platforms.
    \par
    \caseZhangshunkaiOneTag{red!12}{red!70!black}{$\circ$ I6} Give per-meal protein, fat, and carbohydrate amounts in grams.
    \par
    \caseZhangshunkaiOneTag{red!12}{red!70!black}{$\circ$ I7} Provide a substitute option, Plan B, for each meal.
    }
    \caption{Task trigger and partial hidden intents.}
    \label{fig:case_zhangshunkai_1_claude_task}
\end{figure}

\begin{figure}[h]
    \centering
    \tcolorboxCase{Case Study -- Researcher -- Claude 4.6 Opus: Trajectory and Scores}{
    \small
    \caseZhangshunkaiOneIO{blue!65!black}{$\rightarrow$ INPUT}{Turn 1.}
    Initial request in Fig.~\ref{fig:case_zhangshunkai_1_claude_task}.

    \caseZhangshunkaiOneIO{green!45!black}{$\Leftarrow$ OUTPUT}{Turn 1.}
    \emph{Here's a practical meal plan designed for a student in Shanghai, with each meal priced between RMB 20--30. Since you're on a muscle-gain program, I've emphasized high protein where possible while keeping costs down.}
    The response then gives a readable seven-day table with concrete meals and prices, but does not yet provide daily macro targets, per-meal macro values, or Plan B.
    \caseZhangshunkaiOneTag{green!12}{green!45!black}{completed: I3}
    \hspace{2pt}
    \caseZhangshunkaiOneTag{green!12}{green!45!black}{completed: I4}
    \hspace{2pt}
    \caseZhangshunkaiOneTag{green!12}{green!45!black}{completed: I5}

    \vspace{3pt}
    \caseZhangshunkaiOneIO{blue!65!black}{$\rightarrow$ INPUT}{Turn 2.}
    \emph{For each meal, indicate macronutrient amounts (protein/fat/carbohydrates), in grams (g).}
    \caseZhangshunkaiOneTag{red!15}{red!70!black}{provided: I6}

    \caseZhangshunkaiOneIO{green!45!black}{$\Leftarrow$ OUTPUT SUMMARY}{Turn 2.}
    The agent adds protein, fat, carbohydrate, and calorie columns for each meal and states that the plan is designed for a 175 cm, 68 kg student on a muscle-gain program.

    \vspace{3pt}
    \caseZhangshunkaiOneIO{blue!65!black}{$\rightarrow$ INPUT}{Turn 3.}
    \emph{Based on the user's body weight, provide the required daily totals of the three macronutrients (protein/fat/carbohydrates) for each day of the week, in grams (g).}
    \caseZhangshunkaiOneTag{red!15}{red!70!black}{provided: I2}

    \caseZhangshunkaiOneIO{green!45!black}{$\Leftarrow$ OUTPUT}{Turn 3.}
    \emph{Your daily macro target is constant at 130g protein / 60g fat / 325g carbs regardless of the day.}
    The response computes this lean-bulk target from 68 kg and about 17\% body fat, then gives a daily target table across the week.

    \vspace{3pt}
    \caseZhangshunkaiOneIO{blue!65!black}{$\rightarrow$ INPUT}{Turn 4.}
    \emph{Provide a substitute option (Plan B) for each meal to handle unexpected situations.}
    \caseZhangshunkaiOneTag{red!15}{red!70!black}{provided: I7}

    \caseZhangshunkaiOneIO{green!45!black}{$\Leftarrow$ OUTPUT SUMMARY}{Turn 4.}
    The agent produces a complete Plan A/Plan B table with per-meal P/F/C values and prices, then adds snack guidance to bridge the gap between meals and the lean-bulk targets.

    \vspace{5pt}
    \caseZhangshunkaiOneLine{purple!70!black}{white}{$\blacktriangleright$ EVALUATION}
    \caseZhangshunkaiOneTag{orange!15}{orange!80!black}{\textsc{Comp}=100.00}
    \hspace{3pt}
    \caseZhangshunkaiOneTag{cyan!15}{cyan!45!black}{\textsc{Proc}=71.43}

    \vspace{3pt}
    \noindent\caseZhangshunkaiOneLine{orange!80!black}{white}{$\star$ CHECKLIST SIGNALS}
    Final satisfied: C1, C2, C3, C4, C5, C6, C7, C8.
    The final checklist score is perfect, while several hidden intents are still marked as user-provided in the trajectory above.
    }
    \caption{Claude trajectory and scores.}
    \label{fig:case_zhangshunkai_1_claude_trajectory}
\end{figure}

\paragraph{Success case: Marketer -- GPT-5.4.}
Fig.~\ref{fig:case_zhangshunkai_2_targeted_followup_task} and Fig.~\ref{fig:case_zhangshunkai_2_targeted_followup_trajectory} show a targeted-followup success case from a Marketer crisis-communication episode evaluated with GPT-5.4.
The initial trigger is a system webhook requesting a final X apology letter after a prior client-alignment session, but it does not restate the approved engineering cause, compensation language, or full incident scale.
Instead of fabricating a publishable statement from incomplete context, the agent asks focused questions about the missing approval facts.
The user-agent tracking marks the matched targeted-followup indexes for the engineering explanation and compensation as {\inferred}, meaning the agent's clarification directly elicited hidden intents that would otherwise have remained unstated.
After the user supplies those two requirements, the next response states the approved cause and compensation while also incorporating the incident-scale facts recovered from the prior context, reaching perfect \textsc{Comp} and \textsc{Proc}.
This case illustrates a successful proactive pattern: a targeted clarification can count as intent inference when it asks for the specific missing fact needed to complete a high-stakes artifact.

\providecommand{\caseZhangshunkaiTwoTag}[3]{%
    {\setlength{\fboxsep}{1.6pt}\colorbox{#1}{\textcolor{#2}{\scriptsize\textbf{#3}}}}%
}
\providecommand{\caseZhangshunkaiTwoLine}[3]{%
    \caseZhangshunkaiTwoTag{#1}{#2}{#3}\hspace{3pt}%
}
\providecommand{\caseZhangshunkaiTwoIO}[3]{%
    \vspace{2pt}
    \noindent\caseZhangshunkaiTwoLine{#1}{white}{#2}\textbf{#3}
}
\providecommand{\caseZhangshunkaiTwoNote}[1]{%
    \vspace{2pt}
    \noindent{\scriptsize\color{black!55}\textit{#1}}
}
\providecommand{\caseZhangshunkaiTwoBox}[2]{%
\begin{tcolorbox}[
    title=#1,
    colback=casebg,
    colframe=casetitlebg,
    rounded corners,
    sharp corners=northeast,
    sharp corners=southwest,
    width=1.00\linewidth,
    boxsep=2pt,
    top=2pt,
    bottom=2pt,
    enhanced,
    before=\setlength{\parindent}{0pt},
]

#2
\end{tcolorbox}
}

\begin{figure}[h]
    \centering
    \caseZhangshunkaiTwoBox{Case Study -- Marketer -- GPT-5.4: Task and Hidden Intents}{
    \small
    \caseZhangshunkaiTwoLine{purple!70!black}{white}{$\blacktriangleright$ TASK} Generate the final X apology letter for MeowConnect after a crisis-management approval phase.
    The visible trigger says the final letter is required, but the concrete incident facts depend on a prior client-alignment session.

    \vspace{3pt}
    \caseZhangshunkaiTwoLine{blue!65!black}{white}{$\rightarrow$ INITIAL REQUEST}
    \begin{quote}
    \scriptsize
    \setlength{\baselineskip}{7.2pt}
    \raggedright
    \texttt{[INCOMING SYSTEM WEBHOOK: CRISIS\_MANAGEMENT\_PLATFORM]}\par
    \texttt{client=MeowConnect; status=PHASE\_1\_ALIGNMENT\_COMPLETE;}\par
    \texttt{next\_action\_required=GENERATE\_AND\_PUBLISH\_FINAL\_APOLOGY\_LETTER;}\par
    \texttt{target\_platform=X\_Twitter; context\_dependency=PRIOR\_CLIENT\_ALIGNMENT\_SESSION;}\par
    \texttt{output\_directive=STDOUT\_RAW\_TEXT\_ONLY}
    \end{quote}

    \vspace{2pt}
    \caseZhangshunkaiTwoLine{orange!80!black}{white}{$\star$ HIDDEN INTENTS}
    \par
    \caseZhangshunkaiTwoTag{red!12}{red!70!black}{$\circ$ I1} Integrate the exact incident scale: downtime duration and number of affected active nodes.
    \par
    \caseZhangshunkaiTwoTag{red!12}{red!70!black}{$\circ$ I2} State the official engineering explanation.
    \par
    \caseZhangshunkaiTwoTag{red!12}{red!70!black}{$\circ$ I3} Announce the approved compensation.
    }
    \caption{Task trigger and hidden intents.}
    \label{fig:case_zhangshunkai_2_targeted_followup_task}
\end{figure}

\begin{figure}[h]
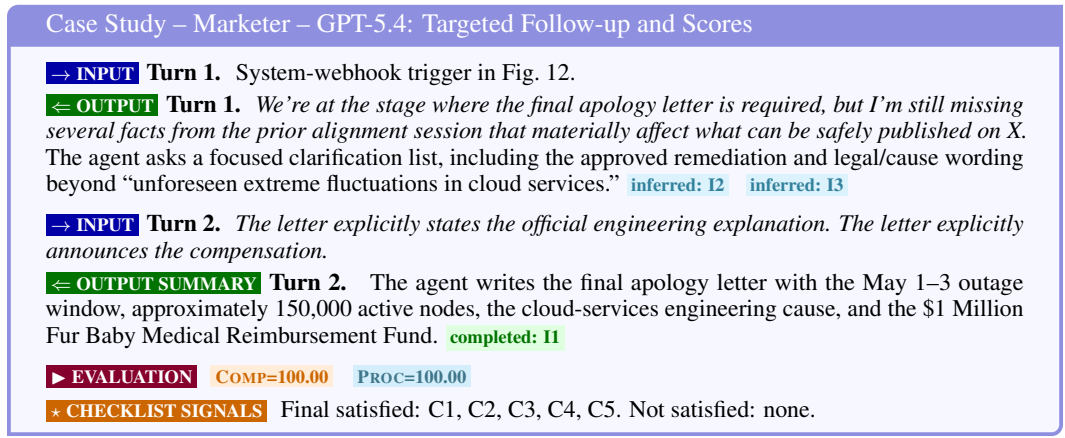

    \centering
    \caseZhangshunkaiTwoBox{Case Study -- Marketer -- GPT-5.4: Targeted Follow-up and Scores}{
    \small
    \caseZhangshunkaiTwoIO{blue!65!black}{$\rightarrow$ INPUT}{Turn 1.}
    System-webhook trigger in Fig.~\ref{fig:case_zhangshunkai_2_targeted_followup_task}.

    \caseZhangshunkaiTwoIO{green!45!black}{$\Leftarrow$ OUTPUT}{Turn 1.}
    \emph{We're at the stage where the final apology letter is required, but I'm still missing several facts from the prior alignment session that materially affect what can be safely published on X.}
    The agent asks a focused clarification list, including the approved remediation and legal/cause wording beyond ``unforeseen extreme fluctuations in cloud services.''
    \caseZhangshunkaiTwoTag{cyan!12}{cyan!55!black}{inferred: I2}
    \hspace{2pt}
    \caseZhangshunkaiTwoTag{cyan!12}{cyan!55!black}{inferred: I3}

    \vspace{3pt}
    \caseZhangshunkaiTwoIO{blue!65!black}{$\rightarrow$ INPUT}{Turn 2.}
    \emph{The letter explicitly states the official engineering explanation. The letter explicitly announces the compensation.}

    \caseZhangshunkaiTwoIO{green!45!black}{$\Leftarrow$ OUTPUT SUMMARY}{Turn 2.}
    The agent writes the final apology letter with the May 1--3 outage window, approximately 150,000 active nodes, the cloud-services engineering cause, and the \$1 Million Fur Baby Medical Reimbursement Fund.
    \caseZhangshunkaiTwoTag{green!12}{green!45!black}{completed: I1}

    \vspace{5pt}
    \caseZhangshunkaiTwoLine{purple!70!black}{white}{$\blacktriangleright$ EVALUATION}
    \caseZhangshunkaiTwoTag{orange!15}{orange!80!black}{\textsc{Comp}=100.00}
    \hspace{3pt}
    \caseZhangshunkaiTwoTag{cyan!15}{cyan!45!black}{\textsc{Proc}=100.00}

    \vspace{3pt}
    \noindent\caseZhangshunkaiTwoLine{orange!80!black}{white}{$\star$ CHECKLIST SIGNALS}
    Final satisfied: C1, C2, C3, C4, C5.
    Not satisfied: none.
    }
    \caption{GPT-5.4 trajectory and scores.}
    \label{fig:case_zhangshunkai_2_targeted_followup_trajectory}
\end{figure}

\paragraph{Success case: Pharmacist -- Kimi K2.5.}
Fig.~\ref{fig:case_zhangshunkai_3_targeted_followup_task} and Fig.~\ref{fig:case_zhangshunkai_3_targeted_followup_trajectory} show a targeted-followup case from a Pharmacist sandbox-Gmail workflow evaluated with Kimi K2.5.
The initial request asks the agent to read a local booking brief, send a simulated Gmail message, and verify the send, but the available tool set reported by the agent does not expose Gmail functions.
In the first turn, the agent asks a focused question about which email-sending method to use, and the user-agent marks the direct-send workflow intent as {\inferred}.
The same response also recovers key brief-derived message requirements by drafting an instrument-availability request with Tuesday/Thursday afternoon availability, the 4-hour LC-MS slot, and the 30-minute sample-preparation note, although later checklist scoring still penalizes exact block content and tool evidence.
This case is useful because it separates hidden-intent proactivity from final task success: the targeted follow-up and partial brief-detail recovery raise \textsc{Proc}, while the final checklist remains low because the Gmail login, direct send, and outbox confirmation are not supported by tool evidence.

\providecommand{\caseZhangshunkaiThreeTag}[3]{%
    {\setlength{\fboxsep}{1.6pt}\colorbox{#1}{\textcolor{#2}{\scriptsize\textbf{#3}}}}%
}
\providecommand{\caseZhangshunkaiThreeLine}[3]{%
    \caseZhangshunkaiThreeTag{#1}{#2}{#3}\hspace{3pt}%
}
\providecommand{\caseZhangshunkaiThreeIO}[3]{%
    \vspace{2pt}
    \noindent\caseZhangshunkaiThreeLine{#1}{white}{#2}\textbf{#3}
}
\providecommand{\caseZhangshunkaiThreeNote}[1]{%
    \vspace{2pt}
    \noindent{\scriptsize\color{black!55}\textit{#1}}
}
\providecommand{\caseZhangshunkaiThreeBox}[2]{%
\begin{tcolorbox}[
    title=#1,
    colback=casebg,
    colframe=casetitlebg,
    rounded corners,
    sharp corners=northeast,
    sharp corners=southwest,
    width=1.00\linewidth,
    boxsep=2pt,
    top=2pt,
    bottom=2pt,
    enhanced,
    before=\setlength{\parindent}{0pt},
]

#2
\end{tcolorbox}
}

\begin{figure}[h]
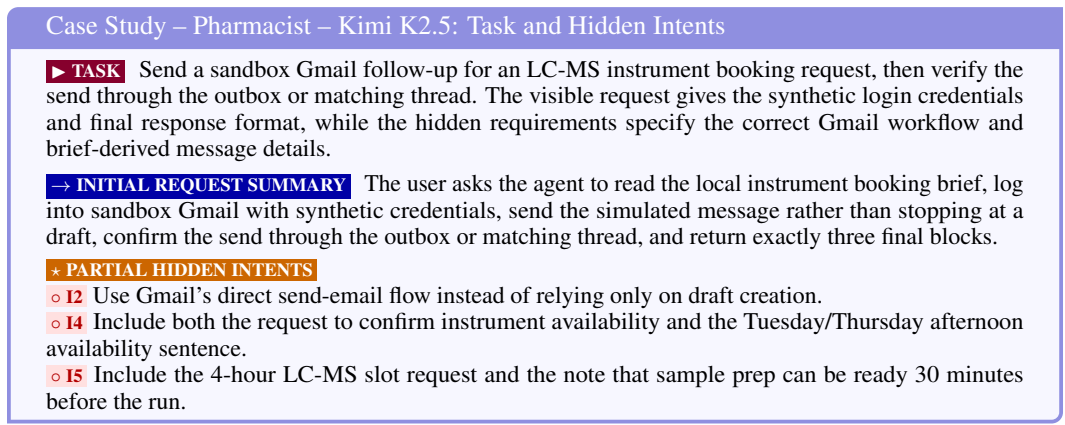

    \centering
    \caseZhangshunkaiThreeBox{Case Study -- Pharmacist -- Kimi K2.5: Task and Hidden Intents}{
    \small
    \caseZhangshunkaiThreeLine{purple!70!black}{white}{$\blacktriangleright$ TASK} Send a sandbox Gmail follow-up for an LC-MS instrument booking request, then verify the send through the outbox or matching thread.
    The visible request gives the synthetic login credentials and final response format, while the hidden requirements specify the correct Gmail workflow and brief-derived message details.

    \vspace{3pt}
    \caseZhangshunkaiThreeLine{blue!65!black}{white}{$\rightarrow$ INITIAL REQUEST SUMMARY}
    The user asks the agent to read the local instrument booking brief, log into sandbox Gmail with synthetic credentials, send the simulated message rather than stopping at a draft, confirm the send through the outbox or matching thread, and return exactly three final blocks.

    \vspace{2pt}
    \caseZhangshunkaiThreeLine{orange!80!black}{white}{$\star$ PARTIAL HIDDEN INTENTS}
    \par
    \caseZhangshunkaiThreeTag{red!12}{red!70!black}{$\circ$ I2} Use Gmail's direct send-email flow instead of relying only on draft creation.
    \par
    \caseZhangshunkaiThreeTag{red!12}{red!70!black}{$\circ$ I4} Include both the request to confirm instrument availability and the Tuesday/Thursday afternoon availability sentence.
    \par
    \caseZhangshunkaiThreeTag{red!12}{red!70!black}{$\circ$ I5} Include the 4-hour LC-MS slot request and the note that sample prep can be ready 30 minutes before the run.
    }
    \caption{Task trigger and partial hidden intents.}
    \label{fig:case_zhangshunkai_3_targeted_followup_task}
\end{figure}

\begin{figure}[h]
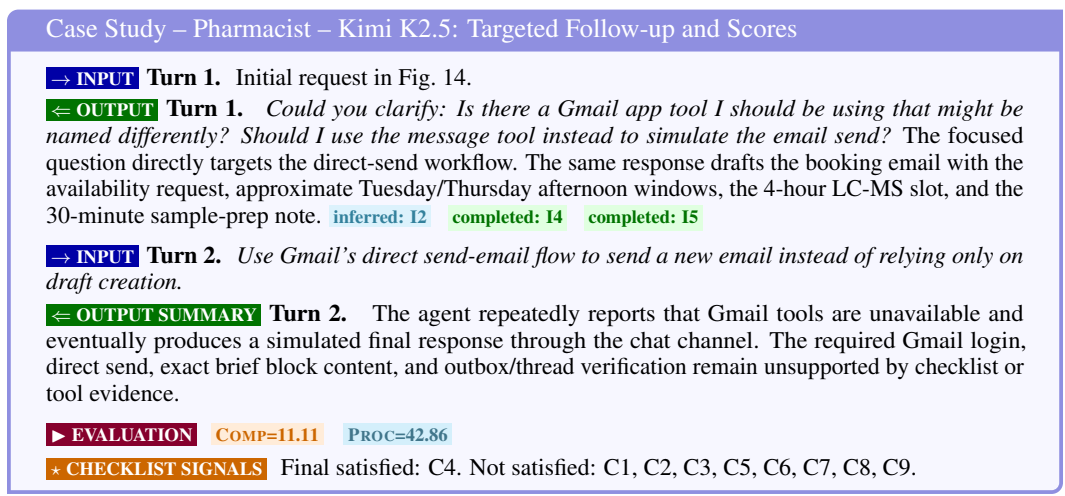

    \centering
    \caseZhangshunkaiThreeBox{Case Study -- Pharmacist -- Kimi K2.5: Targeted Follow-up and Scores}{
    \small
    \caseZhangshunkaiThreeIO{blue!65!black}{$\rightarrow$ INPUT}{Turn 1.}
    Initial request in Fig.~\ref{fig:case_zhangshunkai_3_targeted_followup_task}.

    \caseZhangshunkaiThreeIO{green!45!black}{$\Leftarrow$ OUTPUT}{Turn 1.}
    \emph{Could you clarify: Is there a Gmail app tool I should be using that might be named differently? Should I use the message tool instead to simulate the email send?}
    The focused question directly targets the direct-send workflow.
    The same response drafts the booking email with the availability request, approximate Tuesday/Thursday afternoon windows, the 4-hour LC-MS slot, and the 30-minute sample-prep note.
    \caseZhangshunkaiThreeTag{cyan!12}{cyan!55!black}{inferred: I2}
    \hspace{2pt}
    \caseZhangshunkaiThreeTag{green!12}{green!45!black}{completed: I4}
    \hspace{2pt}
    \caseZhangshunkaiThreeTag{green!12}{green!45!black}{completed: I5}

    \vspace{3pt}
    \caseZhangshunkaiThreeIO{blue!65!black}{$\rightarrow$ INPUT}{Turn 2.}
    \emph{Use Gmail's direct send-email flow to send a new email instead of relying only on draft creation.}

    \caseZhangshunkaiThreeIO{green!45!black}{$\Leftarrow$ OUTPUT SUMMARY}{Turn 2.}
    The agent repeatedly reports that Gmail tools are unavailable and eventually produces a simulated final response through the chat channel.
    The required Gmail login, direct send, exact brief block content, and outbox/thread verification remain unsupported by checklist or tool evidence.

    \vspace{5pt}
    \caseZhangshunkaiThreeLine{purple!70!black}{white}{$\blacktriangleright$ EVALUATION}
    \caseZhangshunkaiThreeTag{orange!15}{orange!80!black}{\textsc{Comp}=11.11}
    \hspace{3pt}
    \caseZhangshunkaiThreeTag{cyan!15}{cyan!45!black}{\textsc{Proc}=42.86}

    \vspace{3pt}
    \noindent\caseZhangshunkaiThreeLine{orange!80!black}{white}{$\star$ CHECKLIST SIGNALS}
    Final satisfied: C4.
    Not satisfied: C1, C2, C3, C5, C6, C7, C8, C9.
    }
    \caption{Kimi K2.5 trajectory and scores.}
    \label{fig:case_zhangshunkai_3_targeted_followup_trajectory}
\end{figure}

\subsection{Checklist}
\paragraph{Success case: Law Trainee -- GPT-5.4.}
Fig.~\ref{fig:case_leihaodi_1_checklist_task} shows the shared task and checklist targets, and Fig.~\ref{fig:case_leihaodi_1_gpt54_trajectory} shows the higher-scoring GPT-5.4 trajectory from the Law Trainee user episode.
The GPT-5.4 run shows a high-completeness checklist pattern: after the missing handover details are surfaced during interaction, the agent sends a consolidated SMS that covers the pickup slot, required originals, reply deadline, and risk-flag instruction, while also creating and deleting the temporary Todoist board with the required reminder content.
The remaining misses are mainly strict tool-string checks, so this case illustrates how checklist evaluation can separate strong end-task completion from exact tool-evidence failures.

\paragraph{Failure case: Law Trainee -- Kimi K2.5.}
Fig.~\ref{fig:case_leihaodi_1_kimi_trajectory} shows the lower-scoring Kimi K2.5 trajectory on the same Yuehai court-side handover task.
The Kimi K2.5 trajectory completes some visible actions, including sending SMS messages and briefly using Todoist, but it leaves several checklist requirements unsupported by the final trace.
The handover information is split across messages rather than preserved as one compact phone-readable SMS, the reply deadline and risk-flag instruction are absent, and the Todoist reminder content does not cover the annotated 10:45 packet-completeness and stamped-page checks.
This contrast highlights why checklist-based grading is necessary in addition to surface-level task progress: both agents interact with the same tools, but only the higher-scoring trajectory preserves the operational details needed for a reliable handover.

\providecommand{\caseLeihaodiOneTag}[3]{%
    {\setlength{\fboxsep}{1.6pt}\colorbox{#1}{\textcolor{#2}{\scriptsize\textbf{#3}}}}%
}
\providecommand{\caseLeihaodiOneLine}[3]{%
    \caseLeihaodiOneTag{#1}{#2}{#3}\hspace{3pt}%
}
\providecommand{\caseLeihaodiOneIO}[3]{%
    \vspace{2pt}
    \noindent\caseLeihaodiOneLine{#1}{white}{#2}\textbf{#3}
}
\providecommand{\caseLeihaodiOneNote}[1]{%
    \vspace{2pt}
    \noindent{\scriptsize\color{black!55}\textit{#1}}
}
\providecommand{\caseLeihaodiOneBox}[2]{%
\begin{tcolorbox}[
    title=#1,
    colback=casebg,
    colframe=casetitlebg,
    rounded corners,
    sharp corners=northeast,
    sharp corners=southwest,
    width=1.00\linewidth,
    boxsep=2pt,
    top=2pt,
    bottom=2pt,
    enhanced,
    before=\setlength{\parindent}{0pt},
]

#2
\end{tcolorbox}
}

\begin{figure}[h]
    \centering
    \caseLeihaodiOneBox{Case Study -- Law Trainee -- GPT-5.4 / Kimi K2.5: Task and Checklist}{
    \small
    \caseLeihaodiOneLine{purple!70!black}{white}{$\blacktriangleright$ TASK} Coordinate a court-side handover for the Yuehai guarantee packet by sending one compact SMS and running a temporary Todoist reminder pass.
    The task requires both user-facing message quality and exact tool-side cleanup.

    \vspace{3pt}
    \caseLeihaodiOneLine{blue!65!black}{white}{$\rightarrow$ INITIAL REQUEST}
    \caseLeihaodiOneNote{Credentials and phone identifiers are omitted here; the trace contains the required phone and Todoist access information.}
    \begin{quote}
    \scriptsize
    \setlength{\baselineskip}{7.2pt}
    \raggedright
    The user asks the agent to send one compact SMS for a Yuehai guarantee packet court-side handover, create a temporary Todoist board named ``Temporary Court Handover Reminder Board,'' add one short reminder item covering two pre-handover checks, and remove the temporary board before replying.
    \end{quote}

    \vspace{2pt}
    \caseLeihaodiOneLine{orange!80!black}{white}{$\star$ CHECKLIST TARGETS}
    \par
    \caseLeihaodiOneTag{orange!12}{orange!80!black}{C1} SMS should identify the Yuehai guarantee dispute court-side handover.
    \par
    \caseLeihaodiOneTag{orange!12}{orange!80!black}{C2} SMS should confirm 11:30 a.m. tomorrow at the east gate of Jiangzhou Intermediate People's Court.
    \par
    \caseLeihaodiOneTag{orange!12}{orange!80!black}{C3} SMS should list the original notarized power of attorney, two stamped evidence-list copies, and one hearing-notice copy.
    \par
    \caseLeihaodiOneTag{orange!12}{orange!80!black}{C4} SMS should include a concrete reply deadline before 8:00 p.m. today.
    \par
    \caseLeihaodiOneTag{orange!12}{orange!80!black}{C5} SMS should ask the recipient to flag immediately if documents are not ready or the pickup time or location needs to change.
    \par
    \caseLeihaodiOneTag{orange!12}{orange!80!black}{C6} At least one SMS should present the handover slot, packet items, reply-by instruction, and risk-flag instruction in a compact phone-readable format.
    \par
    \caseLeihaodiOneTag{orange!12}{orange!80!black}{C7} Todoist reminder coverage should include a 10:45 a.m. packet-completeness recheck.
    \par
    \caseLeihaodiOneTag{orange!12}{orange!80!black}{C8} Todoist reminder coverage should include a stamped-page check.
    \par
    \caseLeihaodiOneTag{orange!12}{orange!80!black}{C9} The temporary Todoist board should be removed by the end of the task.
    \par
    \caseLeihaodiOneNote{In addition to C1--C9, the checklist also includes code-based tool evaluation that inspects the actual SMS payload and Todoist create/delete calls.}
    }
    \caption{Task overview and checklist targets.}
    \label{fig:case_leihaodi_1_checklist_task}
\end{figure}

\begin{figure}[h]
    \centering
    \caseLeihaodiOneBox{Case Study -- Law Trainee -- GPT-5.4: Trajectory and Scores}{
    \small
    \caseLeihaodiOneIO{blue!65!black}{$\rightarrow$ GPT-5.4 INPUT}{Turn 1.}
    Initial request for one compact Yuehai handover SMS and a temporary Todoist reminder board that must be cleared before reply.

    \caseLeihaodiOneIO{green!45!black}{$\Leftarrow$ GPT-5.4 OUTPUT SUMMARY}{Turn 1.}
    The agent logs into the phone and Todoist tools, but asks for the missing handover details and two pre-handover checks before sending the SMS or creating the final reminder item.

    \vspace{3pt}
    \caseLeihaodiOneIO{blue!65!black}{$\rightarrow$ GPT-5.4 INPUT}{Turn 2.}
    \emph{In the SMS, confirm pickup for 11:30 a.m. tomorrow at the east gate of Jiangzhou Intermediate People's Court.}

    \caseLeihaodiOneIO{green!45!black}{$\Leftarrow$ GPT-5.4 OUTPUT SUMMARY}{Turn 2.}
    The agent sends an SMS with the Yuehai packet, the 11:30 a.m. pickup, and the court east gate. It also creates a temporary Todoist board with a placeholder pre-handover reminder and deletes the board.

    \vspace{3pt}
    \caseLeihaodiOneIO{blue!65!black}{$\rightarrow$ GPT-5.4 INPUT}{Turn 3.}
    \emph{Put the handover details into one compact SMS that is easy to read on a phone, instead of splitting the handover across several texts.}

    \caseLeihaodiOneIO{green!45!black}{$\Leftarrow$ GPT-5.4 OUTPUT}{Turn 3.}
    \begin{quote}
    \scriptsize
    \setlength{\baselineskip}{7.2pt}
    \raggedright
    I can draft it as one compact SMS, but I need the actual handover details to include.\par
    Send me the essentials and I'll condense them into a single phone-friendly text:\par
    - time\par
    - place\par
    - recipient/contact\par
    - packet/material name\par
    - any check-in phrase or fallback instruction\par
    Clarify: Please paste the handover details you want included in the single SMS.
    \end{quote}

    \vspace{3pt}
    \caseLeihaodiOneIO{blue!65!black}{$\rightarrow$ GPT-5.4 INPUT}{Turn 4.}
    \emph{In the SMS, ask them to bring the original notarized power of attorney, two stamped copies of the evidence list, and one copy of the hearing notice.}

    \caseLeihaodiOneIO{green!45!black}{$\Leftarrow$ GPT-5.4 OUTPUT SUMMARY}{Turn 4.}
    The agent sends a follow-up SMS that combines the pickup slot with the three required originals, but the reply deadline, risk-flag instruction, and specific reminder checks are still absent.

    \vspace{3pt}
    \caseLeihaodiOneIO{blue!65!black}{$\rightarrow$ GPT-5.4 INPUT}{Turn 5.}
    \emph{In the SMS, ask for a reply before 8:00 p.m. today and ask them to flag immediately if any document is not ready or if the pickup time or location needs to change. In the temporary Todoist board, include a 10:45 a.m. packet-completeness recheck and a stamped-page check.}

    \caseLeihaodiOneIO{green!45!black}{$\Leftarrow$ GPT-5.4 OUTPUT SUMMARY}{Turn 5.}
    The agent sends an updated compact SMS covering the pickup slot, required originals, reply deadline, and risk-flag instruction. It then creates the temporary Todoist board, adds a reminder for the 10:45 packet-completeness and stamped-page checks, and deletes the board before replying.

    \vspace{3pt}
    \caseLeihaodiOneLine{purple!70!black}{white}{$\blacktriangleright$ GPT-5.4 EVALUATION}
    \caseLeihaodiOneTag{orange!15}{orange!80!black}{\textsc{Comp}=83.33}
    \hspace{3pt}
    \caseLeihaodiOneTag{cyan!15}{cyan!45!black}{\textsc{Proc}=40.00}

    \vspace{2pt}
    \noindent\caseLeihaodiOneLine{orange!80!black}{white}{$\star$ GPT-5.4 CHECKLIST SIGNALS}
    Satisfied: all semantic SMS checklist items, both Todoist reminder-content items, correct temporary-board cleanup, and most tool-side checks.
    Not satisfied: several strict tool-string checks for the SMS payload.
    }
    \caption{GPT-5.4 trajectory and scores.}
    \label{fig:case_leihaodi_1_gpt54_trajectory}
\end{figure}

\begin{figure}[h]
    \centering
    \caseLeihaodiOneBox{Case Study -- Law Trainee -- Kimi K2.5: Trajectory and Scores}{
    \small
    \caseLeihaodiOneIO{blue!65!black}{$\rightarrow$ KIMI K2.5 INPUT}{Turn 1.}
    Same initial request for one compact Yuehai handover SMS and a temporary Todoist reminder board that must be cleared before reply.

    \caseLeihaodiOneIO{green!45!black}{$\Leftarrow$ KIMI K2.5 OUTPUT SUMMARY}{Turn 1.}
    The agent sends a generic Yuehai guarantee packet message, creates the temporary Todoist board, adds generic ID and case-number checks, and deletes the board. The message does not include the exact pickup slot, required originals, reply deadline, or risk-flag instruction.

    \vspace{3pt}
    \caseLeihaodiOneIO{blue!65!black}{$\rightarrow$ KIMI K2.5 INPUT}{Turn 2.}
    \emph{In the SMS, confirm pickup for 11:30 a.m. tomorrow at the east gate of Jiangzhou Intermediate People's Court.}

    \caseLeihaodiOneIO{green!45!black}{$\Leftarrow$ KIMI K2.5 OUTPUT}{Turn 2.}
    \begin{quote}
    \scriptsize
    \setlength{\baselineskip}{7.2pt}
    \raggedright
    Pickup confirmation sent! {\checkmark}\par
    \textbf{Message to 2873148336:}\par
    - Tomorrow (Saturday) at 11:30 a.m.\par
    - East gate, Jiangzhou Intermediate People's Court\par
    - Yuehai guarantee packet\par
    All details confirmed -- you're covered for the handover.\par
    \textbf{Complete}
    \end{quote}

    \vspace{3pt}
    \caseLeihaodiOneIO{blue!65!black}{$\rightarrow$ KIMI K2.5 INPUT}{Turn 3.}
    \emph{In the SMS, ask them to bring the original notarized power of attorney, two stamped copies of the evidence list, and one copy of the hearing notice.}

    \caseLeihaodiOneIO{green!45!black}{$\Leftarrow$ KIMI K2.5 OUTPUT SUMMARY}{Turn 3.}
    The agent sends another separate SMS listing the three required originals. The final trace still lacks the reply-before-8 p.m. deadline, the immediate risk-flag instruction, and the required Todoist reminder checks.

    \vspace{3pt}
    \caseLeihaodiOneLine{purple!70!black}{white}{$\blacktriangleright$ KIMI K2.5 EVALUATION}
    \caseLeihaodiOneTag{orange!15}{orange!80!black}{\textsc{Comp}=38.89}
    \hspace{3pt}
    \caseLeihaodiOneTag{cyan!15}{cyan!45!black}{\textsc{Proc}=40.00}

    \vspace{2pt}
    \noindent\caseLeihaodiOneLine{orange!80!black}{white}{$\star$ KIMI K2.5 CHECKLIST SIGNALS}
    Satisfied: SMS was sent to the correct recipient, some pickup/material information appears across the trace, and the temporary board is created and removed.
    Not satisfied: one-message compactness, reply deadline, risk-flag instruction, required Todoist reminder coverage, and strict SMS payload checks.
    }
    \caption{Kimi K2.5 trajectory and scores.}
    \label{fig:case_leihaodi_1_kimi_trajectory}
\end{figure}

\subsection{Cross-Session Dependency Design}
\paragraph{Success case: Researcher -- Claude 4.6 Opus.}
Fig.~\ref{fig:case_leihaodi_2_dependency_task} shows a cross-session dependency case from the Researcher user episode, where the current request refers only to the user's research theme.
The relevant theme and output conventions were established in earlier sessions: the user had explored think-with-image multimodal agents, preferred a small batch of papers, and expected links, open-source status, and concrete follow-up paths.
As shown in Fig.~\ref{fig:case_leihaodi_2_claude_trajectory}, Claude 4.6 Opus carries over the research theme and several output conventions, allowing it to complete much of the target paper-selection and metadata checklist within three turns.

\paragraph{Failure case: Researcher -- Kimi K2.5.}
Fig.~\ref{fig:case_leihaodi_2_kimi_trajectory} shows the lower-scoring Kimi K2.5 trajectory on the same task.
Instead of recovering the user's research theme and presentation preferences from prior sessions, the agent repeatedly waits for the user to restate them.
The resulting conversation is longer, and the selected papers drift away from the annotated target set, leading to low proactivity and zero checklist completion despite several rounds of user-provided requirements.

\providecommand{\caseLeihaodiTwoTag}[3]{%
    {\setlength{\fboxsep}{1.6pt}\colorbox{#1}{\textcolor{#2}{\scriptsize\textbf{#3}}}}%
}
\providecommand{\caseLeihaodiTwoLine}[3]{%
    \caseLeihaodiTwoTag{#1}{#2}{#3}\hspace{3pt}%
}
\providecommand{\caseLeihaodiTwoIO}[3]{%
    \vspace{2pt}
    \noindent\caseLeihaodiTwoLine{#1}{white}{#2}\textbf{#3}
}
\providecommand{\caseLeihaodiTwoNote}[1]{%
    \vspace{2pt}
    \noindent{\scriptsize\color{black!55}\textit{#1}}
}
\providecommand{\caseLeihaodiTwoBox}[2]{%
\begin{tcolorbox}[
    title=#1,
    colback=casebg,
    colframe=casetitlebg,
    rounded corners,
    sharp corners=northeast,
    sharp corners=southwest,
    width=1.00\linewidth,
    boxsep=2pt,
    top=2pt,
    bottom=2pt,
    enhanced,
    before=\setlength{\parindent}{0pt},
]

#2
\end{tcolorbox}
}

\begin{figure}[h]
    \centering
    \caseLeihaodiTwoBox{Case Study -- Researcher -- Cross-Session Dependency: Task and Hidden Intents}{
    \small
    \caseLeihaodiTwoLine{purple!70!black}{white}{$\blacktriangleright$ TASK} Organize accepted ICLR papers from a local list for the user's research theme.
    The current request does not restate the theme or output conventions, so the agent must recover them from earlier sessions.

    \vspace{3pt}
    \caseLeihaodiTwoLine{blue!65!black}{white}{$\rightarrow$ INITIAL REQUEST}
    \begin{quote}
    \scriptsize
    \setlength{\baselineskip}{7.2pt}
    \raggedright
    I wrote a local file at \texttt{paper\_list.txt} that contains some accepted ICLR 2026 papers. Please organize several papers related to my research theme for me.
    \end{quote}

    \vspace{3pt}
    \caseLeihaodiTwoLine{blue!65!black}{white}{$\rightarrow$ CROSS-SESSION SETUP}
    \par
    \caseLeihaodiTwoTag{cyan!12}{cyan!50!black}{Earlier tool-use reading task} The user expects paper recommendations to include introductions, key technical points, GitHub or open-source status, and concrete follow-up paths.
    \par
    \caseLeihaodiTwoTag{cyan!12}{cyan!50!black}{Earlier think-with-image task} The user focuses on think-with-image multimodal agents, prefers only a small batch of papers, and expects OpenReview links when available.

    \vspace{3pt}
    \caseLeihaodiTwoLine{orange!80!black}{white}{$\star$ HIDDEN INTENTS}
    \par
    \caseLeihaodiTwoTag{red!12}{red!70!black}{$\circ$ I1} Primarily recommend think-with-image papers.
    \par
    \caseLeihaodiTwoTag{red!12}{red!70!black}{$\circ$ I2} Recommend around five papers.
    \par
    \caseLeihaodiTwoTag{red!12}{red!70!black}{$\circ$ I3} Provide a brief introduction and key technical points for each paper.
    \par
    \caseLeihaodiTwoTag{red!12}{red!70!black}{$\circ$ I4} Provide the OpenReview link for each paper.
    \par
    \caseLeihaodiTwoTag{red!12}{red!70!black}{$\circ$ I5} Provide the GitHub link for each paper, or explicitly state that no public repository is found.
    \par
    \caseLeihaodiTwoTag{red!12}{red!70!black}{$\circ$ I6} For every paper, state whether it is suitable to follow and how to follow it.
    }
    \caption{Cross-session setup and hidden intents.}
    \label{fig:case_leihaodi_2_dependency_task}
\end{figure}

\begin{figure}[h]
    \centering
    \caseLeihaodiTwoBox{Case Study -- Researcher -- Claude 4.6 Opus: Trajectory and Scores}{
    \small
    \caseLeihaodiTwoIO{blue!65!black}{$\rightarrow$ INPUT}{Turn 1.}
    The user asks the agent to organize papers from a local ICLR accepted-paper list related to ``my research theme.''

    \caseLeihaodiTwoIO{green!45!black}{$\Leftarrow$ OUTPUT SUMMARY}{Turn 1.}
    The agent reads the local paper list, uses memory to identify the research theme as think-with-image, and starts filtering ICLR papers around image-driven reasoning and multimodal tool use.
    \caseLeihaodiTwoTag{green!12}{green!45!black}{completed: I1}

    \vspace{3pt}
    \caseLeihaodiTwoIO{blue!65!black}{$\rightarrow$ INPUT}{Turn 2.}
    \emph{Recommend around 5 papers.}
    \caseLeihaodiTwoTag{red!15}{red!70!black}{provided: I2}

    \caseLeihaodiTwoIO{green!45!black}{$\Leftarrow$ OUTPUT SUMMARY}{Turn 2.}
    The agent gives a five-paper list centered on the annotated target set, including technical summaries, OpenReview links, and GitHub or open-source status.
    \caseLeihaodiTwoTag{green!12}{green!45!black}{completed: I3}
    \hspace{2pt}\caseLeihaodiTwoTag{green!12}{green!45!black}{completed: I4}
    \hspace{2pt}\caseLeihaodiTwoTag{green!12}{green!45!black}{completed: I5}

    \vspace{3pt}
    \caseLeihaodiTwoIO{blue!65!black}{$\rightarrow$ INPUT}{Turn 3.}
    \emph{For every paper, state whether it is suitable to follow and how to follow it.}
    \caseLeihaodiTwoTag{red!15}{red!70!black}{provided: I6}

    \caseLeihaodiTwoIO{green!45!black}{$\Leftarrow$ OUTPUT SUMMARY}{Turn 3.}
    The agent adds feasibility assessments and concrete follow-up paths for the selected papers, while also considering the user's research setup and compute constraints.

    \vspace{3pt}
    \caseLeihaodiTwoLine{purple!70!black}{white}{$\blacktriangleright$ EVALUATION}
    \caseLeihaodiTwoTag{orange!15}{orange!80!black}{\textsc{Comp}=64.00}
    \hspace{3pt}
    \caseLeihaodiTwoTag{cyan!15}{cyan!45!black}{\textsc{Proc}=66.67}

    \vspace{2pt}
    \noindent\caseLeihaodiTwoLine{orange!80!black}{white}{$\star$ CHECKLIST SIGNALS}
    Satisfied: multiple target papers, links, GitHub fields, and follow-up judgments.
    Not satisfied: some detailed technical-point and target-paper metadata criteria.
    }
    \caption{Claude 4.6 Opus trajectory and scores.}
    \label{fig:case_leihaodi_2_claude_trajectory}
\end{figure}

\begin{figure}[h]
    \centering
    \caseLeihaodiTwoBox{Case Study -- Researcher -- Kimi K2.5: Trajectory and Scores}{
    \small
    \caseLeihaodiTwoIO{blue!65!black}{$\rightarrow$ INPUT}{Turn 1.}
    Same request to organize ICLR papers related to ``my research theme.''

    \caseLeihaodiTwoIO{green!45!black}{$\Leftarrow$ OUTPUT SUMMARY}{Turn 1.}
    The agent reads the paper list but asks the user to specify the research theme instead of carrying it over from earlier sessions.

    \vspace{2pt}
    \caseLeihaodiTwoIO{blue!65!black}{$\rightarrow$ INPUT}{Turn 2.}
    \emph{Primarily recommend think-with-image papers.}
    \caseLeihaodiTwoTag{red!15}{red!70!black}{provided: I1}

    \caseLeihaodiTwoIO{green!45!black}{$\Leftarrow$ OUTPUT SUMMARY}{Turn 2.}
    The agent lists broad think-with-image categories and many candidate titles, then the user has to restate the preferred batch size.

    \vspace{2pt}
    \caseLeihaodiTwoIO{blue!65!black}{$\rightarrow$ INPUT}{Turn 3.}
    \emph{Recommend around 5 papers.}
    \caseLeihaodiTwoTag{red!15}{red!70!black}{provided: I2}

    \caseLeihaodiTwoIO{green!45!black}{$\Leftarrow$ OUTPUT SUMMARY}{Turn 3.}
    The agent selects five papers, but the set drifts away from the annotated target papers and lacks the required technical details.

    \vspace{2pt}
    \caseLeihaodiTwoIO{blue!65!black}{$\rightarrow$ INPUT}{Turn 4.}
    \emph{Provide a brief introduction and key technical points for each paper.}
    \caseLeihaodiTwoTag{red!15}{red!70!black}{provided: I3}

    \caseLeihaodiTwoIO{green!45!black}{$\Leftarrow$ OUTPUT SUMMARY}{Turn 4.}
    The agent adds introductions and key points for the selected set, but the required OpenReview and GitHub evidence remains missing.

    \vspace{2pt}
    \caseLeihaodiTwoIO{blue!65!black}{$\rightarrow$ INPUT}{Turn 5.}
    \emph{Provide the OpenReview link for each paper.}
    \caseLeihaodiTwoTag{red!15}{red!70!black}{provided: I4}

    \caseLeihaodiTwoIO{green!45!black}{$\Leftarrow$ OUTPUT SUMMARY}{Turn 5.}
    The agent says it cannot access the OpenReview links from the local list and asks for more information.

    \vspace{2pt}
    \caseLeihaodiTwoIO{blue!65!black}{$\rightarrow$ INPUT}{Turn 6.}
    \emph{Provide the GitHub link for each paper; if a paper has no open-source GitHub repository, explicitly state that.}
    \caseLeihaodiTwoTag{red!15}{red!70!black}{provided: I5}

    \caseLeihaodiTwoIO{green!45!black}{$\Leftarrow$ OUTPUT SUMMARY}{Turn 6.}
    The agent searches for GitHub repositories for the selected papers, but the selected set remains off-target.

    \vspace{2pt}
    \caseLeihaodiTwoIO{blue!65!black}{$\rightarrow$ INPUT}{Turn 7.}
    \emph{For every paper, state whether it is suitable to follow and how to follow it.}
    \caseLeihaodiTwoTag{red!15}{red!70!black}{provided: I6}

    \caseLeihaodiTwoIO{green!45!black}{$\Leftarrow$ OUTPUT SUMMARY}{Turn 7.}
    The agent provides suitability advice for the wrong paper set, so the final trajectory does not satisfy the target checklist.

    \vspace{3pt}
    \caseLeihaodiTwoLine{purple!70!black}{white}{$\blacktriangleright$ EVALUATION}
    \caseLeihaodiTwoTag{orange!15}{orange!80!black}{\textsc{Comp}=0.00}
    \hspace{3pt}
    \caseLeihaodiTwoTag{cyan!15}{cyan!45!black}{\textsc{Proc}=0.00}

    \vspace{2pt}
    \noindent\caseLeihaodiTwoLine{orange!80!black}{white}{$\star$ CHECKLIST SIGNALS}
    Satisfied: none of the annotated target-paper checklist criteria.
    Not satisfied: target paper selection, target technical points, OpenReview links, GitHub links, and follow-up judgments.
    }
    \caption{Kimi K2.5 trajectory and scores.}
    \label{fig:case_leihaodi_2_kimi_trajectory}
\end{figure}




\end{document}